\documentclass{article} % For LaTeX2e

\usepackage[svgnames,usenames,dvipsnames,table]{xcolor}         % colors
\usepackage{amsmath,amsfonts,bm}

\def\eqref#1{equation~\ref{#1}}
\def\1{\bm{1}}

\DeclareMathAlphabet{\mathsfit}{\encodingdefault}{\sfdefault}{m}{sl}
\SetMathAlphabet{\mathsfit}{bold}{\encodingdefault}{\sfdefault}{bx}{n}

\newcommand{\E}{\mathbb{E}}
\newcommand{\Ls}{\mathcal{L}}

\usepackage{booktabs}       % professional-quality tables
\usepackage[hyphens]{url}
\usepackage[colorlinks=true, citecolor=Navy, breaklinks=true]{hyperref}       % hyperlinks
\usepackage{nicefrac}       % compact symbols for 1/2, etc.
\usepackage{caption}
\usepackage{subcaption}
\usepackage{graphicx}
\usepackage{wrapfig}
\usepackage[most]{tcolorbox}
\usepackage{multirow}
\usepackage[capitalize,noabbrev]{cleveref}
\usepackage{enumitem}
\usepackage{cereb}

\usepackage{tikz}
\usepackage{pgfplots}
\usetikzlibrary{arrows.meta}  % Load the arrows.meta library

\newcommand{\onefigscale}{0.55}
\newcommand{\wdfigscales}{0.93}
\newcommand{\shrinkfigtwo}{0.45}
\newcommand{\shrinkfigimgs}{0.49}

\newcommand{\hateta}{\tilde{\eta}}
\newcommand{\maxlr}{1.6\mbox{e-}02}
\newcommand{\maxlrdetail}{1.62\mbox{e-}02}
\newcommand{\tenx}{\mbox{10}\times}
\newcommand{\dtoz}{\mbox{D2Z}}
\newcommand{\mup}{\mbox{$\mu$P}}
\newcommand{\supar}{\mbox{S$\mu$Par}}

\newcommand{\invsqrt}{\mbox{\emph{InvSqrt}}}
\newcommand{\constant}{\mbox{\emph{Constant}}}
\newcommand{\linear}{\mbox{\emph{Linear}}}
\newcommand{\cosine}{\mbox{\emph{Cosine}}}
\newcommand{\rational}{\mbox{\emph{Rational}}}
\newcommand{\step}{\mbox{\emph{Step}}}
\newcommand{\wsd}{\mbox{\emph{WSD}}}
\newcommand{\cyclic}{\mbox{\emph{Cyclic}}}
\newcommand{\baralpha}{\bar{\alpha}}
\newcommand{\dmodel}{d_{\text{model}}}
\newcommand{\dhead}{d_{\text{head}}}
\newcommand{\dffn}{d_{\text{ffn}}}
\newcommand{\nlayers}{n_{\text{layers}}}

\newcommand{\tikzscale}{1.2}
\newcommand{\tikztextscale}{0.75}

\tcbset{textmarker/.style={%
  enhanced,
  parbox=false,boxrule=0mm,boxsep=0mm,arc=0mm,
  outer arc=0mm,left=6mm,right=3mm,top=4pt,bottom=3pt,
  toptitle=1mm,bottomtitle=1mm,oversize,
  left skip=0mm,
  right skip=8mm
}}

\newcounter{fcounter}
\newcommand\finding[1]{
        \refstepcounter{fcounter}\vspace{2pt}
        \begin{tcolorbox}[colback=yellow!10!white,colframe=yellow!80!black,boxsep=1pt,left=2pt,right=2pt,top=1pt,bottom=1pt]\noindent{\textbf{\sffamily Finding \arabic{fcounter}}: \sffamily #1}
        \end{tcolorbox}\vspace{0pt}
}
\crefname{fcounter}{Finding}{Findings}

\newcounter{kcounter}
\crefname{kcounter}{Takeaway}{Takeaways}

\newtcolorbox{hypothesisBox}{textmarker,
    borderline west={6pt}{0pt}{blue},
    colback=blue!10!white}
\newcounter{hcounter}
\newcommand\hypothesis[1]{
        \refstepcounter{hcounter}\vspace{1.0pt}
        \begin{hypothesisBox}\noindent{\textbf{\sffamily Hypothesis \arabic{hcounter}}: \sffamily #1}
        \end{hypothesisBox}\vspace{-0.5pt}
}
\crefname{hcounter}{Hypothesis}{Hypotheses}

\renewcommand{\cite}[1]{\PackageError{MyPackage}{Do not use \string\cite\space with natbib. Use \string\citet\space or \string\citep}{See the natbib package documentation for explanation.}}

\makeatletter
\AtBeginDocument{%
  \@ifpackageloaded{cleveref}{%

    \providecommand\cref@appendix@setup{%
      \crefname{appendix}{Appendix}{Appendices}%
      \Crefname{appendix}{Appendix}{Appendices}%
    }%

    \AddToHook{cmd/appendix/before}{%
      \cref@appendix@setup
      \@ifundefined{chapter}{%
        \crefalias{section}{appendix}%
      }{%
        \crefalias{chapter}{appendix}%
      }%
      \crefalias{subsection}{appendix}%
      \crefalias{subsubsection}{appendix}%
    }%

  }{}% end \@ifpackageloaded{cleveref}
}
\makeatother

\title{Straight to Zero: Why Linearly Decaying the Learning Rate to Zero Works Best for LLMs}

\author{Shane Bergsma, Nolan Dey, Gurpreet Gosal, Gavia Gray, Daria Soboleva \& Joel Hestness \\
  Cerebras Systems \\
  \texttt{\{shane.bergsma,joel\}@cerebras.net}
}

\begin{document}

\typeout{Topmargin=\the\topmargin, Headheight=\the\headheight,
  Headsep=\the\headsep, Textheight=\the\textheight, Footskip=\the\footskip}

\maketitle
\vspace{-2mm}

\typeout{Topmargin=\the\topmargin, Headheight=\the\headheight,
  Headsep=\the\headsep, Textheight=\the\textheight, Footskip=\the\footskip}

\begin{abstract}
%Many different learning rate (LR) schedules have been used to train
%LLMs.
%
%With little understanding of which schedules work best in which
%situations,
%
LLMs are commonly trained with a learning rate (LR) warmup, followed
by cosine decay to 10\% of the maximum ($\tenx$ decay).
In a large-scale empirical study, we show that under an optimal peak
LR, a simple linear decay-to-zero ($\dtoz$) schedule consistently
outperforms other schedules when training at compute-optimal dataset
sizes.
$\dtoz$ is superior across a range of model sizes, batch sizes,
datasets, and vocabularies.  Benefits increase as dataset size
increases.
Leveraging a novel interpretation of AdamW as an exponential moving
average of weight updates, we show how linear $\dtoz$ optimally
balances the demands of early training (moving away from initial
conditions) and late training (averaging over more updates in order to
mitigate gradient noise).
In experiments, a 610M-parameter model trained for 80
tokens-per-parameter (TPP) using $\dtoz$ achieves \emph{lower} loss
than when trained for 200~TPP using $\tenx$ decay, corresponding to an
astonishing 60\% compute savings.
Models such as Llama2-7B, trained for 286~TPP with $\tenx$ decay,
could likely have saved a majority of compute by training with
$\dtoz$.
All the main experiments were run on Cerebras CS-3 systems.

%experimental settings. These observations prove robust
%across model sizes, batch sizes, architectures, weight-sparsity
%levels, and parameterizations.

%We explain the success of linear $\dtoz$ by interpreting AdamW as a
%convex combination of weight updates, with combination coefficients a
%function of the LR schedule; linear $\dtoz$ is best able to balance
%effective early training (combining a narrow window of updates in
%order to forget initial conditions) and reducing gradient noise later
%on (smoothing over many updates).

\end{abstract}

\vspace{-3mm}

\section{Introduction}\label{sec:intro}
Learning rate (LR) schedules play an important role in training large
language models.  The original Transformers paper proposed a brief LR
warmup followed by decay proportional to the inverse square root of
the step number~\citep{vaswani2017attention}.
%
%This schedule, sometimes
%called \emph{NoamOpt}~\citep{rush2018annotated} and often used in work
%by Noam Shazeer and
%colleagues~\citep{shazeer2017outrageously,raffel2020exploring,lepikhin2020gshard,zoph2022st,chowdhery2022palm},
%
This schedule has the advantage of not requiring prior specification
of the total training steps.  However, cooling down to a specific
minimum LR is acknowledged to be ``preferable when one knows the
training duration in advance''~\citep{zhai2022scaling} as it produces
``slightly better results''~\citep{raffel2020exploring}.
In this paper, our main focus is finding LR schedules that achieve the
minimum loss given a pre-specified number of training tokens.

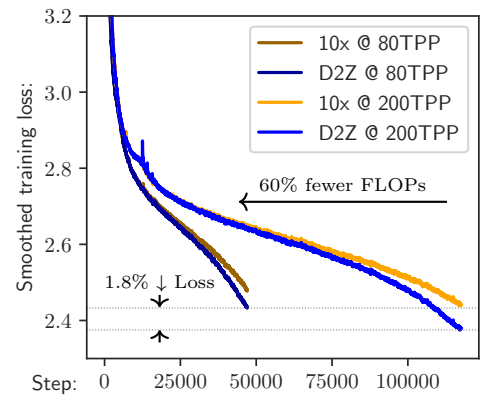
\begin{wrapfigure}{r}{0.38\textwidth}
  \centering
  \vspace{-5mm}
  \begin{tikzpicture}
    % Place the image in a node
    \node[anchor=south west,inner sep=0] (image) at (0,0) {
      \scalebox{0.39}{
        \includegraphics[trim={0.35cm 0cm 0cm 0.3cm}, clip, width=\textwidth]{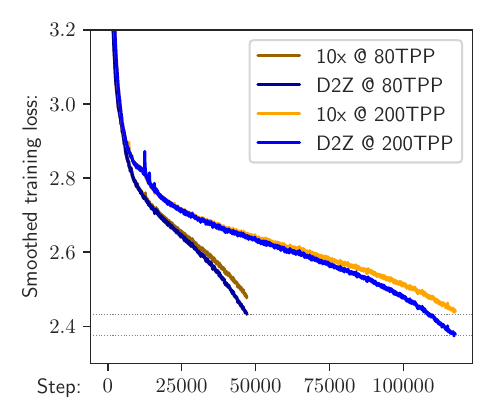}
      }
    };
    % Set up a coordinate system relative to the image
    \begin{scope}[x={(image.south east)},y={(image.north west)}]
      % FLOP savings line
      \draw[<-,thick] (0.46,0.52) -- (0.881,0.52)
      node[midway, above] {\scriptsize 60\% fewer FLOPs};
      % Vertical arrowed lines
      \draw[->,thick] (0.3,0.295) -- (0.3,0.265)
      node[midway, above] {\scriptsize 1.8\% $\downarrow$ Loss};
      \draw[->,thick] (0.3,0.18) -- (0.3,0.21)
      node[midway] {};
      % Add text above the top arrow
      % \node at (0.3, 0.3) {\scriptsize 1.8\%};
    \end{scope}
  \end{tikzpicture}
  \mbox{}
  \vspace{-7mm}
  \mbox{}
  \caption{\small A 610M model trained for 80~TPP with
    $\linear$-$\dtoz$ has better train (and \emph{validation}) loss
    than when trained for 200~TPP with $\linear$-$\tenx$.
    %The gap
    %between $\tenx$ and $\dtoz$ increases with
    %TPP.
    \label{fig:train_loss}}
  \vspace{-2mm}
\end{wrapfigure}

The ``predominant choice''~\citep{hu2024minicpm} in such
training---the ``de-facto standard''~\citep{hagele2024scaling}---is
warmup followed by cosine decay to 10\% of the peak LR, an approach
used in GPT3~\citep{brown2020language}, Gopher~\citep{rae2022scaling},
Chinchilla~\citep{hoffmann2022empirical},
\textsc{bloom}~\citep{lescao2023bloom}, Llama~\citep{touvron2023llama},
Llama2~\citep{touvron2023llama2}, Falcon~\citep{almazrouei2023falcon},
Pythia~\citep{biderman2023pythia}, etc.
It is used ``following \citeauthor{hoffmann2022empirical}''~\citep{muennighoff2023scaling}, and is the default in
LLM codebases~\citep{karpathy2024nanogpt}.
% NanoGPT: ``minimum learning rate, should be ~= learning\_rate/10 per
% Chinchilla''

We present a large-scale empirical study to determine which schedules
work best in which situations, and why.
We focus on both \emph{compute-efficient} and \emph{over-trained}
models.  According to Chinchilla scaling
laws \citep{hoffmann2022empirical}, the fewest FLOPs to achieve a
given loss is obtained when models are trained for around 20
tokens-per-parameter (TPP).  It is also common to train for more than
20~TPP because smaller, over-trained models are cheaper to
serve \citep{touvron2023llama}.
% We hypothesized that the optimal LR schedule may depend on the peak
% LR, and validated this empirically.
Our experiments (across various model scales, vocabulary sizes, and
dataset sources) reveal a consistent outcome: when all schedules use
their optimal peak LR, linear decay-to-zero ($\dtoz$) works best at
compute-optimal TPP\@.
Moreover, the relative benefit of $\dtoz$ over $\tenx$ (in terms of
training, validation and downstream loss) \emph{increases} with TPP\@.
Compute savings from $\dtoz$ can be substantial for over-trained
models (\cref{fig:train_loss}).
LLMs such as Llama2-7B~\citep{touvron2023llama2}, trained for 286~TPP
at $\tenx$ decay, could likely have saved most of their training
compute by switching to $\dtoz$.

\begin{figure}
  \centering
  \makebox[\textwidth][c]{
    \begin{subfigure}{\shrinkfigtwo\textwidth}
      \includegraphics[trim={0.1cm 0.1cm 0.1cm 0.2cm}, clip, width=\textwidth]{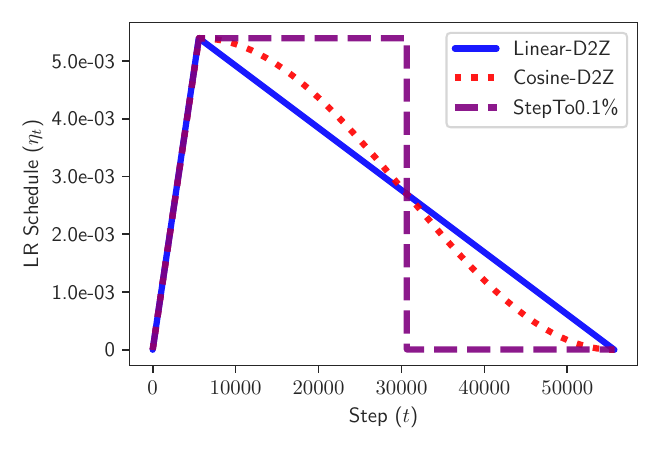}
    \end{subfigure}
    \hspace{-1mm}
    \begin{subfigure}{\shrinkfigtwo\textwidth}
      \includegraphics[trim={0.1cm 0.1cm 0.1cm 0.2cm}, clip, width=\textwidth]{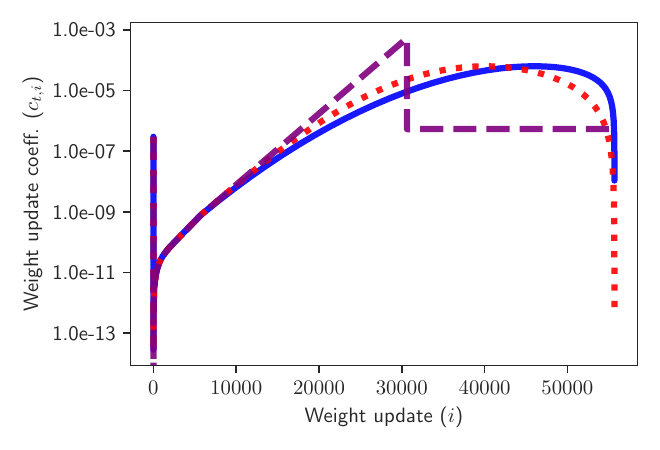}
    \end{subfigure}
  }
  \mbox{}
  \vspace{-7mm}
  \mbox{}
  \caption{\textbf{LR schedules and their update-combination duals}:
    Each LR schedule, $\eta_t$ (left) and weight decay, $\lambda$,
    implies a weighted combination of weight \emph{updates}, with
    combination coefficients $c_{t,i}$ (right, log-scale) giving the
    contribution of $i$th update to parameters $\theta_t$ at step $t$
    (111M scale, coefficients at final step).
    The more sudden the drop in LR, the less emphasis on valuable
    later updates, perhaps explaining why $\step$ underperforms
    $\cosine$ and $\cosine$ underperforms $\linear$
    decay.\label{fig:background_ema}
    %    corresponding to settings for 111M-param $\mup$ models:
    %    $\hateta$=$\maxlr$, $\rho$=$\nicefrac{1}{3}$, $\lambda$=$0.1$.
  }
\end{figure}

To explain the success of $\linear$-$\dtoz$, we build on recent work
on the related topic of \emph{weight
decay}~\citep{andriushchenko2023why,wang2024how}.
First, decaying \emph{to zero} helps because compute-efficient
training includes a long phase in which gradient noise is the key
factor slowing the loss reduction; with higher noise resulting from
higher TPP \emph{or} smaller batches, a vanishing LR works best.
% result in less emphasis on recent weight updates, and,
%consequentially, worse model quality.
Second, we show that approaching zero \emph{linearly} is beneficial
via a novel interpretation of AdamW~\citep{loshchilov2017decoupled},
the primary optimizer in LLM pre-training.
% -- as a convex combination of weight updates
With AdamW, weights generated at each step are implicitly a weighted
combination of weight \emph{updates}.  The combination coefficients
depend on the learning rate schedule and weight decay settings
(\cref{subsec:convex}).
Analyzing this \emph{dual} of the LR schedule, we observe linear decay
to produce a favorable combination of prior weight updates
(\cref{fig:background_ema}).  When LR drops abruptly, e.g., via
step-decay or, to a lesser extent, cosine decay, later updates receive
less emphasis, leading to worse model quality.
The EMA dual perspective also reveals the implicit schedule-awareness
of recent ``schedule-free'' approaches such
as \emph{Warmup-Stable-Decay}
(WSD)~\citep{hu2024minicpm,bi2024deepseek,hagele2024scaling}.
%However, it also suggests a method for truly schedule-free training
%(\cref{subsec:rational}).
%
%Using the dual view of the LR schedule to interpret our experimental
%results, we develop a multi-part theory for the success of $\dtoz$.

Our empirical study encompasses hundreds of models trained over a grid
of schedules and peak LRs, with scales ranging from 111M to 1.7B
parameters, and datasets up to 137B tokens.  We also compare LR
schedules over the cross product of weight decay versus peak LR, and
peak LR versus batch size.
We confirm a slight, consistent advantage of linear $\dtoz$ over
cosine $\dtoz$ at 20~TPP\@.  We also demonstrate that linear $\dtoz$
improves over continuous schedules such as WSD\@.
Further, when dataset or batch size changes, optimal peak LRs are much
more \emph{stable} when using $\dtoz$ compared to using lesser decay.
This latter finding exposes LR decay ratio as an important confounder
in prior work studying optimal hyperparameter transfer with
$\mup$~\citep{yang2020feature,yang2022mup}.

\section{Background and Related Work}\label{sec:background}
\subsection{Learning rate schedules}\label{subsec:lr_schedules}

LR schedules have a long history in stochastic optimization, and are
motivated by convergence bounds for stochastic gradient
methods~\citep{moulines2011non,bottou2018optimization}.  For example,
following \citet{andriushchenko2023why}, consider SGD for a convex
loss parameterized by $\theta$: with a constant LR $\eta$, the gap
between the optimum and current loss at step $t$ can be bounded by:
\begin{equation}\label{eqn:biasvar}
\E[\Ls(\theta_t) - \Ls(\theta_*)] \le
\textcolor{TealBlue}{(1 - \eta\mu)^t ||\theta_0 - \theta_*||^2 }
+ \textcolor{orange}{\eta \sigma^2}
\end{equation}
where $\theta_0$ are initial parameters, $\theta_*$ is the loss
minimizer, $\sigma^2$ is a bound on the variance of gradient noise,
and $\mu$ is a measure of the objective's curvature (see also
\citet[Theorem~4.6]{bottou2018optimization}).  A larger LR can
decrease dependence on initial conditions (the
\textcolor{TealBlue}{\emph{bias}} term), but also increase the effect
of gradient noise (the \textcolor{orange}{\emph{variance}} term).
As training progresses, bias decreases exponentially in $t$, and the
relative importance of variance increases.
\citet{bottou2018optimization} note this motivates a strategy where LR
is high initially (to mitigate bias) and lowered later (to minimize
variance).

In practice, when and how to reduce the LR is rarely informed by ML
theory.
Many LLMs simply follow the 10x cosine schedule, which is noted to
work slightly better than cosine $\dtoz$ in the influential work of
\citet{hoffmann2022empirical}.
It is also well established that using a portion of a longer (or
extending a shorter) schedule is suboptimal compared to using a
schedule that reaches its minimum only at the final training
step~\citep{li2019budgeted,hoffmann2022empirical,hu2024minicpm,hagele2024scaling}.
%
%\citet{hoffmann2022empirical} adopt cosine decay, and report that
%``the difference between decaying by 10 and decaying to 0.0 ... is
%small, though decaying by a factor of 10 to be slightly more
%performant,'' while ``decaying by less (5) is clearly worse.''
%
Linear decay after warmup~\citep{howard2018universal} has been used in
LLMs, typically also to 10\% of the
peak~\citep{henighan2020scaling,dey2023cerebras,sengupta2023jais}.
Dropping LR at specific milestones (\emph{step decay}) is popular in
vision models~\citep{he2016deep,zagoruyko2016wide,li2020reconciling},
but has also been used in LLMs~\citep{bi2024deepseek}.
% do drop the LR in stages.  Seems to be popularized with
% \citet{zagoruyko2016wide}.  Although \citet{andriushchenko2023why}
% (who use it) cite ``step-decay as LR schedule'' from .  ``Step
% Decay, one of the most commonly-used learning rate
% schedules''~\citep{li2020reconciling}

\citet{kaplan2020scaling} compared various decay functions and
concluded the specific schedule was unimportant given a high enough
average LR, although decaying to zero ``gives a fixed improvement
close to the end of training.''
%They also use $\dtoz$ in \citet{radford2018improving}.
%Aside from the $\dtoz$ bonus, they say it really just seems to depend
%on the total area, but their results are noisy.  In our work, we
%consistently find, e.g., Cosine is worse than Linear.
%
Few papers explicitly compare different LR schedules for large-scale
training, and when comparisons are
made~\citep{shallue2019measuring,kaplan2020scaling,schmidt2021descending,hoffmann2022empirical,yang2022mup},
they are not the primary focus.  So, while some insights are gained,
``comprehensive study'' is usually regarded as ``out of
scope''~\citep{alephalpha2024introducing}.
% Could be a footnote: this situation is not unique to LR schedules;
% given the huge costs of LLM training at
% scale~\citep{strubell2019energy,patterson2021carbon,bender2021dangers}
% "There is little room for hyperparameter tuning." (Du et al, Glam
% paper).
%
One exception is \citet{defazio2023when}, who found linear equals or
outperforms other common schedules, including cosine, across a range
of problems, including LLM training.  In addition,
\citeauthor{defazio2023when}\ develop convergence bounds that
theoretically motivate linear as the optimal schedule.  Unlike our
work, they do not evaluate LLMs with different peak LRs or decay
ratios.

Seeking to measure model quality at different training durations
without having to re-train separate models from
scratch~\citep{zhai2022scaling,hagele2024scaling}, researchers have
adopted various \emph{continuous} schedules, such as constant,
cyclic~\citep{smith2017cyclical}, etc.
Following optimization theory~\citep{moulines2011non,defazio2024road}
weight averaging provides an alternative to
decay~\citep{sandler2023training,sanyal2023early,busbridge2024scale},
although it is typically not as effective~\citep{hagele2024scaling},
and moreover may have hyperparameters that implicitly depend on
training duration~\citep{defazio2024road}.
\emph{Warmup-Stable-Decay} (WSD) approaches have also been used in LLM
training~\citep{hu2024minicpm,shen2024jetmoe,ibrahim2024simple,hagele2024scaling}.
These methods train at a constant LR, but decay from a checkpoint in a
separate process when an intermediate model is needed.
We show that the optimal constant LR of these methods implicitly
depends on training duration, rendering them not truly
\emph{schedule-free}.

\subsection{Maximal update parameterization ($\mup$)}\label{subsec:mup}

It is common to scale initial weights of neural networks such that
activations have unit
variance~\citep{glorot2010initialization,he2015delving}.  Yet weights
can become unstable after a few steps of updates, if layer-wise LRs
are imbalanced~\citep{yang2022mup}.
% Instabilities grow with width~\citep{yang2022mup}.
In contrast, $\mup$~\citep{yang2020feature,dey2024practitioner}
prescribes a \emph{re-parameterization} of initial weight variances
and LRs -- essentially, rules for scaling these values as model width
(i.e., $d_{model}$) scales -- so that activations and updates remain
stable.
$\mup$ also stabilizes embeddings, layer norms, and self-attention in
Transformers.
%~\cite[App. B.1]{yang2022mup}.
% It enables \emph{feature learning} (separation of concepts in
%embedding space) at the infinite-width limit, while such learning
%notably does not occur with the standard or
%NTK~\citep{jacot2018neural}
%parameterizations~\citep{yang2020feature,noci2024learning}.  The
%ability of $\mup$ to facilitate HP transfer was developed
%in~\citep{yang2022mup} and a spectral perspective is provided
%in~\citep{yang2023spectral}.
%
$\mup$ is seeing growing application in
LLMs~\citep{dey2023cerebras,shen2024power,hu2024minicpm}, where it
acts to stabilize training and to enable transfer of optimal
hyperparameters (HPs) across model scales.

With $\mup$, base HPs can be tuned on a small \emph{proxy}, or
\emph{base}, model, and transferred to larger models.  Given the width
of the proxy model, $d_p$, and width of the target, $d_t$, $\mup$
prescribes scaling factors to apply to HPs. The base LR $\hateta$ is
scaled down to $\eta = \rho\hateta$, where $\rho=\nicefrac{d_p}{d_t}$.
In terms of LR schedules, the base LR $\hateta_t$ is scaled at every
step to provide $\eta_t$.  $\mup$ is convenient in our study as we can
sweep the same \emph{base} peak LRs, $\hateta$, at each model size,
and observe trends that are scale-invariant.

\subsection{AdamW weights as an exponentially-weighted moving average (EMA)}\label{subsec:wang}

An AdamW update at a single training step, $t$, can be expressed as:
\begin{equation}
\theta_t = (1 - \eta\lambda)\theta_{t-1} - \eta \frac{\hat{m}_t}{\sqrt{\hat{v}_t} + \epsilon}\label{eqn:adamw}
\end{equation}
where $\eta$ is the learning rate, $\hat{m}_t$ and $\hat{v}_t$ are
(bias-corrected) running averages of the gradient and the squared
gradient, respectively, and $\epsilon$ is a small constant added to
prevent division by zero.  The weight decay value, $\lambda$, is
typically set to 0.1 in LLM
training~\citep{brown2020language,hoffmann2022empirical,almazrouei2023falcon,alephalpha2024introducing}.

While the running averages of $m$ and $v$ in
Adam~\citep{kingma2014adam} and AdamW are exponentially-weighted
moving averages (EMAs), \citet{wang2024how} recently showed that the
\emph{weights} generated by AdamW can also be understood as an EMA ---
of the weight \emph{updates}.  That is, a standard EMA, $y_t$, for a
time-varying quantity, $x_t$, can be written as:
\begin{equation}\label{eqn:ema}
y_t = (1 - \alpha)y_{t-1} + \alpha x_t
\end{equation}
where $\alpha$ is the \emph{smoothing parameter}.  AdamW in
\cref{eqn:adamw} can be seen as an EMA by letting:
\begin{equation}\label{eqn:adamwema}
y_t=\theta_t\mbox{, }\alpha=\eta\lambda\mbox{, and }
x_t=-\frac{1}{\lambda}\frac{\hat{m}_t}{\sqrt{\hat{v}_t} + \epsilon}
\end{equation}
\citeauthor{wang2024how} note the quantity
$\tau$=$\nicefrac{1}{\alpha}$, i.e., $\nicefrac{1}{(\eta\lambda)}$,
provides a rough \emph{timescale} over which updates are averaged.
Weight decay can therefore be used to control the effective window
over which updates are combined (smaller values of $\lambda$ increase
$\tau$, increasing the role of earlier updates in $\theta_t$).
This perspective also motivates dynamic LR schedules.  Initial $\tau$
should be small (high $\eta_t$), to forget early updates, ``while the
final timescale is around the total number of epochs [low $\eta_t$],
to ensure averaging over all datapoints.''
In fact, we will show that contribution of weight updates, $x_0, x_1,
\ldots x_{t}$, to $\theta_t$ at a particular step, $t$, cannot be
determined by the \emph{instantaneous} value of $\eta_t\lambda$.
Rather, the contribution of any $x_i$ to $\theta_t$ requires looking
at the full LR schedule \emph{holistically} (\cref{subsec:convex}).

\citeauthor{wang2024how} also motivate scaling rules for optimal
$\lambda$ as model and dataset size vary.  If dataset size increases
by a factor of $K$ (i.e., $K\times$ as many steps), the EMA
perspective recommends scaling $\lambda$ by $\nicefrac{1}{K}$ in order
to expand $\tau$ proportional to $K$.
Moreover, with $\mup$, if model size increases and LR is scaled by
$\rho$ (\cref{subsec:mup}), the EMA perspective motivates scaling
$\lambda$ by $\nicefrac{1}{\rho}$ to keep $\tau$
constant.\footnote{While the EMA perspective does not account for the
updates $x_t$ themselves depending on parameters $y_{t-1}$, we find it
a useful part of our \emph{conceptual} model of training, in that it
helps predict experimental results.}
%
%Although not explored by \citeauthor{wang2024how}, analagous rules can
%be derived for batch size changes.
%
%Of course, in practice we typically scale batch and dataset size
%together with model size, so real application requires jointly
%accounting for multiple factors.

\section{Methods, Conceptual Foundations, and Hypotheses}\label{sec:emas}
We now present an extended EMA perspective on AdamW that accounts for
time-varying LRs.  We then introduce our conceptual model of LLM
training, and connect it to the EMA perspective.  Finally, we outline
the specific testable hypotheses that follow from our conceptual
model.

\subsection{AdamW as convex combination of weight updates, driven by LR schedule}\label{subsec:convex}

Consider a generic moving average, but now with time-varying
smoothing, $\alpha_t$, i.e., $y_t = (1 - \alpha_t)y_{t-1} + \alpha_t
x_t$.  If we let $\alpha_1=1$ (so that $y_1=x_1$), we can express
$y_t$ in terms of all inputs $x_t$:
\begin{eqnarray}\label{eqn:extended_ema}
  y_1 &=& \alpha_1 x_1  \notag, \\
  y_2 &=& (1 - \alpha_2) \alpha_1 x_1 + \alpha_2 x_2, \cdots  \notag \\
%  y_3 &=& (1 - \alpha_3) (1 - \alpha_2) \alpha_1 x_1 + (1 - \alpha_3) \alpha_2 x_2 + \alpha_3 x_3, \cdots \notag \\
%  &\cdots&  \notag \\
  y_t &=& \sum_{i=1}^t \left( \prod_{j=i+1}^{t} (1 - \alpha_j) \right) \alpha_i x_i \notag \\
      &=& \sum_{i=1}^t c_{t,i} x_i
\end{eqnarray}

Where $c_{t,i} = \left( \prod_{j=i+1}^{t} (1 - \alpha_j) \right) \alpha_i$.
%
%is the contribution of input $x_i$ to output $y_t$ at time $t$.
%such
%that
%
%$y_t = \sum_{i=1}^t c_{t,i} x_i$.
%
It can be shown
$\forall t, \sum_i c_{t,i} = 1$, and hence, as in a standard EMA, each
$y_t$ is a \emph{convex combination} of inputs $x_1 \ldots x_t$ (all
coefficients are non-negative and sum to 1).  However, with
time-varying $\alpha_t$, we are not restricted to
exponentially-decreasing $c_{t,i}$ as $i$ decreases.  Indeed, for
\emph{any} convex combination of inputs, there is a corresponding
smoothing schedule that generates the combination via the EMA\@.
In terms of LR schedules for AdamW training, $y_t = \theta_t$, while
$\alpha_t = \eta_t\lambda$ becomes the smoothing parameter at step $t$
(cf. \cref{eqn:ema,eqn:adamwema}).\footnote{When using a $\mup$ LR
scaling factor, $\rho$ (\cref{subsec:mup}), the smoothing parameter is
$\alpha = \eta\lambda = \rho\hateta\lambda$.}

\subsection{Conceptual model: bias and variance in LLM training}\label{subsec:biasvar}

\begin{figure}[th]
\centering

\begin{subfigure}[t]{0.32\textwidth}
  \centering
\begin{tikzpicture}[scale=\tikzscale, every node/.style={scale=\tikztextscale}]
    % Define the dimensions of the bounding box
    \def\boxwidth{2.2} % Width for a wider appearance
    \def\rectwidth{1.5}
    \def\rectwidthTwo{0.7}
    \def\boxheight{1.5} % Less height compared to the width
    \def\rectheight{0.7}
    \def\rectheightTwo{0.7} % A different height for the second rectangle
    \def\schedshift{0.8}

    % Points for the line near the edges of rectangles
    \coordinate (LineStart) at (0.3, \rectheightTwo/2 + \schedshift); % 10pt from left edge
    \coordinate (LineEnd) at (\boxwidth - 0.3, \rectheightTwo/2 + \schedshift); % 10pt from right edge

    % Draw the rectangle in the upper-left quadrant with very light green fill
    \fill[green!20] (0, \boxheight - \rectheight) rectangle (\rectwidth, \boxheight);

    % Draw the rectangle in the bottom-right quadrant with very light red fill, different height
    \fill[red!20] (\boxwidth - \rectwidthTwo, 0) rectangle (\boxwidth, \rectheightTwo);

    % Draw the bounding box
    \draw (0, 0) rectangle (\boxwidth, \boxheight);

    % Draw a line between the modified points near the edges of the rectangles
    \draw[{Circle[length=4mm, width=4mm, color=blue]}-{Circle[length=4mm, width=4mm, color=blue]}, line width=2pt, blue] (LineStart) -- (LineEnd);

    \node[fill=white,draw=blue!80,fill opacity=.90] at ([xshift=3.5pt,yshift=-5pt]LineStart) {maxLR};
    \node[fill=white,draw=blue!80,fill opacity=.90] at ([xshift=-3pt,yshift=-5pt]LineEnd) {minLR};

    % Labels for Bias and Variance reduction
    \node[align=center,draw=green!30] (BiasLabel) at (\rectwidth/2, \boxheight + 0.2) {Bias reduction};
    \node[align=center,draw=red!30] (VarLabel) at (\boxwidth - \rectwidthTwo/2, \rectheightTwo - 0.2) {Variance reduction};
    
    % Thin lines from labels to rectangles
    \draw[thin] (BiasLabel.south) -- (\rectwidth/2 + 0.05, \boxheight - 0.05);
    \draw[thin] (VarLabel.south) -- (\boxwidth - \rectwidthTwo/2 + 0.05, \rectheightTwo -0.2 - 0.2 - 0.05);

    % Axis labels
    \node at (\boxwidth/2, -0.2) {Optimization steps};  % X-axis label
    \node[rotate=90] at (-0.2, \boxheight/2) {LR setting};  % Y-axis label

    % Arrows on axis labels
    \draw[-Stealth] (\boxwidth - 0.2, -0.2) -- (\boxwidth - 0.0, -0.2);
    \draw[-Stealth] (-0.2, \boxheight - 0.2) -- (-0.2, \boxheight - 0.0);
\end{tikzpicture}
\vspace{-0.1cm}
\caption{$\constant$, fewer steps (low TPP)\label{fig:const_low}}
\end{subfigure}%
\hfill
\begin{subfigure}[t]{0.65\textwidth}
  \centering
\begin{tikzpicture}[scale=\tikzscale, every node/.style={scale=\tikztextscale}]
    % Define the dimensions of the bounding box
    \def\boxwidth{6}
    \def\rectwidth{1.5}
    \def\rectwidthTwo{4.5}
    \def\boxheight{1.5}
    \def\rectheight{0.7}
    \def\rectheightTwo{0.7}
    \def\schedshift{0.2}

    % Points for the line near the edges of rectangles
    \coordinate (LineStart) at (0.3, \rectheightTwo/2 + \schedshift); % 10pt from left edge
    \coordinate (LineEnd) at (\boxwidth - 0.3, \rectheightTwo/2 + \schedshift); % 10pt from right edge

    % Draw the rectangle in the upper-left quadrant with very light green fill
    \fill[green!20] (0, \boxheight - \rectheight) rectangle (\rectwidth, \boxheight);

    % Draw the rectangle in the bottom-right quadrant with very light red fill, different height
    \fill[red!20] (\boxwidth - \rectwidthTwo, 0) rectangle (\boxwidth, \rectheightTwo);

    % Draw the bounding box
    \draw (0, 0) rectangle (\boxwidth, \boxheight);

    % Draw a line between the modified points near the edges of the rectangles
    \draw[{Circle[length=4mm, width=4mm, color=blue]}-{Circle[length=4mm, width=4mm, color=blue]}, line width=2pt, blue] (LineStart) -- (LineEnd);

    \node[fill=white,draw=blue!80,fill opacity=.90] at ([xshift=3.5pt,yshift=-5pt]LineStart) {maxLR};
    \node[fill=white,draw=blue!80,fill opacity=.90] at ([xshift=-3pt,yshift=-5pt]LineEnd) {minLR};

    % Labels for Bias and Variance reduction
    \node[align=center,draw=green!30] (BiasLabel) at (\rectwidth/2, \boxheight + 0.2) {Bias reduction};
    \node[align=center,draw=red!30] (VarLabel) at (\boxwidth - \rectwidthTwo/2, \rectheightTwo + 0.2) {Variance reduction};
    
    % Thin lines from labels to rectangles
    \draw[thin] (BiasLabel.south) -- (\rectwidth/2 + 0.05, \boxheight - 0.05);
    \draw[thin] (VarLabel.south) -- (\boxwidth - \rectwidthTwo/2 + 0.05, \rectheightTwo - 0.05);

    % Axis labels
    \node at (\boxwidth/2, -0.2) {Optimization steps};  % X-axis label
    \node[rotate=90] at (-0.2, \boxheight/2) {LR setting};  % Y-axis label

    % Arrows on axis labels
    \draw[-Stealth] (\boxwidth - 2.1, -0.2) -- (\boxwidth - 1.9, -0.2);
    \draw[-Stealth] (-0.2, \boxheight - 0.2) -- (-0.2, \boxheight - 0.0);
\end{tikzpicture}
\vspace{-0.1cm}
\caption{$\constant$, many steps (high TPP)\label{fig:const_high}}
\end{subfigure}

\vspace{0.2cm} % Space between the rows

\noindent

\begin{subfigure}[t]{0.32\textwidth}
  \centering
\begin{tikzpicture}[scale=\tikzscale, every node/.style={scale=\tikztextscale}]
    % Define the dimensions of the bounding box
    \def\boxwidth{2.2} % Width for a wider appearance
    \def\rectwidth{1.5}
    \def\rectwidthTwo{0.7}
    \def\boxheight{1.5} % Less height compared to the width
    \def\rectheight{0.7}
    \def\rectheightTwo{0.7} % A different height for the second rectangle
    \def\schedshift{0.17}
    \def\schedshiftTwo{-0.25}

    % Points for the line near the edges of rectangles
    \coordinate (LineStart) at (0.3,  \boxheight - \rectheight/2 + \schedshift); % 10pt from left edge
    \coordinate (LineEnd) at (\boxwidth - 0.3, \rectheightTwo/2 + \schedshiftTwo); % 10pt from right edge

    % Draw the rectangle in the upper-left quadrant with very light green fill
    \fill[green!20] (0, \boxheight - \rectheight) rectangle (\rectwidth, \boxheight);

    % Draw the rectangle in the bottom-right quadrant with very light red fill, different height
    \fill[red!20] (\boxwidth - \rectwidthTwo, 0) rectangle (\boxwidth, \rectheightTwo);

    % Draw the bounding box
    \draw (0, 0) rectangle (\boxwidth, \boxheight);

    % Draw a line between the modified points near the edges of the rectangles
    \draw[{Circle[length=4mm, width=4mm, color=blue]}-{Circle[length=4mm, width=4mm, color=blue]}, line width=2pt, blue] (LineStart) -- (LineEnd);

    \node[fill=white,draw=blue!80,fill opacity=.90] at ([xshift=3.5pt,yshift=-5pt]LineStart) {maxLR};
    \node[fill=white,draw=blue!80,fill opacity=.90] at ([xshift=-3pt,yshift=-1pt]LineEnd) {minLR};

    % Labels for Bias and Variance reduction
    \node[align=center,draw=green!30] (BiasLabel) at (\rectwidth/2, \boxheight + 0.2) {Bias reduction};
    \node[align=center,draw=red!30] (VarLabel) at (\boxwidth - \rectwidthTwo/2, \rectheightTwo + 0.2) {Variance reduction};
    
    % Thin lines from labels to rectangles
    \draw[thin] (BiasLabel.south) -- (\rectwidth/2 + 0.05, \boxheight - 0.05);
    \draw[thin] (VarLabel.south) -- (\boxwidth - \rectwidthTwo/2 + 0.05, \rectheightTwo - 0.05);

    % Axis labels
    \node at (\boxwidth/2, -0.2) {Optimization steps};  % X-axis label
    \node[rotate=90] at (-0.2, \boxheight/2) {LR setting};  % Y-axis label

    % Arrows on axis labels
    \draw[-Stealth] (\boxwidth - 0.2, -0.2) -- (\boxwidth - 0.0, -0.2);
    \draw[-Stealth] (-0.2, \boxheight - 0.2) -- (-0.2, \boxheight - 0.0);
\end{tikzpicture}
\vspace{-0.1cm}
\caption{$\dtoz$, fewer steps (low TPP)\label{fig:dtoz_low}}
\end{subfigure}%
\hfill
\begin{subfigure}[t]{0.65\textwidth}
  \centering
\begin{tikzpicture}[scale=\tikzscale, every node/.style={scale=\tikztextscale}]
    % Define the dimensions of the bounding box
    \def\boxwidth{6}
    \def\rectwidth{1.5}
    \def\rectwidthTwo{4.5}
    \def\boxheight{1.5}
    \def\rectheight{0.7}
    \def\rectheightTwo{0.7}
    \def\schedshift{-0.05}
    \def\schedshifttwo{-0.25}

    % Points for the line near the edges of rectangles
    \coordinate (LineStart) at (0.3, \boxheight - \rectheight/2 + \schedshift);
    \coordinate (LineEnd) at (\boxwidth - 0.3, \rectheightTwo/2 + \schedshifttwo);

    % Draw the rectangle in the upper-left quadrant with very light green fill
    \fill[green!20] (0, \boxheight - \rectheight) rectangle (\rectwidth, \boxheight);

    % Draw the rectangle in the bottom-right quadrant with very light red fill, different height
    \fill[red!20] (\boxwidth - \rectwidthTwo, 0) rectangle (\boxwidth, \rectheightTwo);

    % Draw the bounding box
    \draw (0, 0) rectangle (\boxwidth, \boxheight);

    % Draw a line between the modified points near the edges of the rectangles
    \draw[{Circle[length=4mm, width=4mm, color=blue]}-{Circle[length=4mm, width=4mm, color=blue]}, line width=2pt, blue] (LineStart) -- (LineEnd);

    \node[fill=white,draw=blue!80,fill opacity=.90] at ([xshift=3.5pt,yshift=-5pt]LineStart) {maxLR};
    \node[fill=white,draw=blue!80,fill opacity=.90] at ([xshift=-3pt,yshift=-5pt]LineEnd) {minLR};

    % Labels for Bias and Variance reduction
    \node[align=center,draw=green!30] (BiasLabel) at (\rectwidth/2, \boxheight + 0.2) {Bias reduction};
    \node[align=center,draw=red!30] (VarLabel) at (\boxwidth - \rectwidthTwo/2, \rectheightTwo + 0.2) {Variance reduction};
    
    % Thin lines from labels to rectangles
    \draw[thin] (BiasLabel.south) -- (\rectwidth/2 + 0.05, \boxheight - 0.05);
    \draw[thin] (VarLabel.south) -- (\boxwidth - \rectwidthTwo/2 + 0.05, \rectheightTwo - 0.05);

    % Axis labels
    \node at (\boxwidth/2, -0.2) {Optimization steps};  % X-axis label
    \node[rotate=90] at (-0.2, \boxheight/2) {LR setting};  % Y-axis label
    
    % Arrows on axis labels
    \draw[-Stealth] (\boxwidth - 2.1, -0.2) -- (\boxwidth - 1.9, -0.2);
    \draw[-Stealth] (-0.2, \boxheight - 0.2) -- (-0.2, \boxheight - 0.0);
\end{tikzpicture}
\vspace{-0.1cm}
\caption{$\dtoz$, many steps (high TPP)\label{fig:dtoz_high}}
\end{subfigure}

\caption{\textbf{Bias \& variance in LLM pre-training}: as training
  duration increases (higher TPP), the importance of variance
  reduction --- and having a lower LR --- increases.  With no decay
  ($\constant$, \ref{fig:const_low}, \ref{fig:const_high}), the
  optimal peak LR must drop significantly lower, neglecting bias
  reduction.  With $\dtoz$ (\ref{fig:dtoz_low}, \ref{fig:dtoz_high}),
  periods of bias and variance reduction can both be enjoyed without
  large shifts in peak LR.\label{fig:bias_var_all}}
\end{figure}
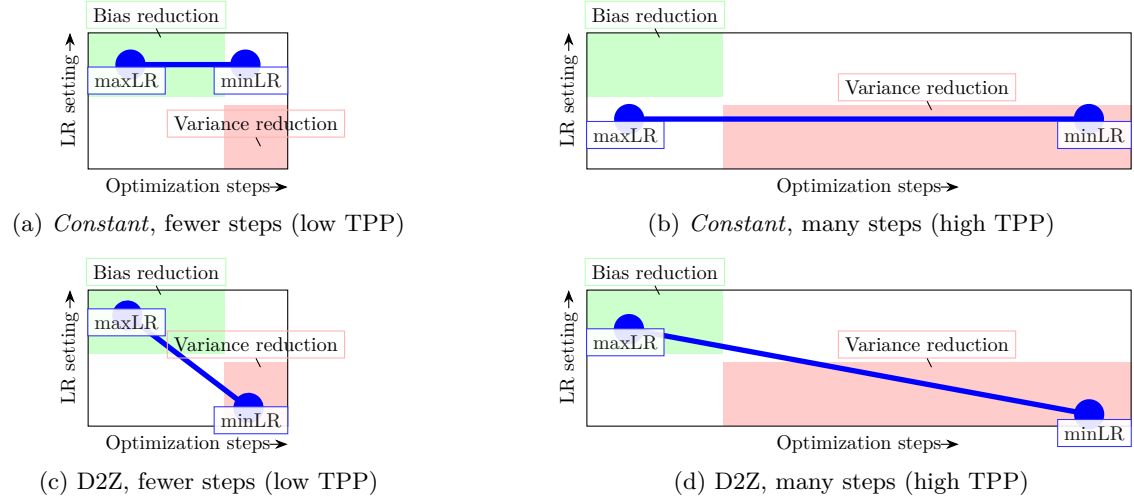

Following \citet{andriushchenko2023why}, our main conceptual premise
is that in LLM training, there is an initial training phase focused on
movement from initial conditions (bias reduction), followed by a later
phase bottlenecked by gradient variance.
Furthermore, analogous to prior work optimizing the convex loss gap
via SGD (\cref{eqn:biasvar}), we argue the primary beneficial
mechanism of LR decay in LLM training is to reduce gradient noise
during later stages of training, while simultaneously minimizing the
impact on bias reduction.
The larger the dataset, the longer the proportion of training
bottlenecked by gradient noise, and the greater the benefit from $\dtoz$
(\cref{fig:bias_var_all}).

\paragraph{Variance reduction}

\begin{wrapfigure}{r}{0.33\textwidth}
\centering
\vspace{-9mm}
\begin{tikzpicture}[scale=0.76]
    \def\growthxspikept{1.7}
    \def\growthyflatpt{0.02}
    \def\growthdescale{10}
    \begin{axis}[
        axis lines=middle,
        xlabel={TPP},
        ymin=0, ymax=0.6,
        xmin=0, xmax=10,
        width=1.6\linewidth,  % Set width relative to the wrapfigure width
        height=1.0\linewidth, % Set height to maintain aspect ratio
        x label style={at={(axis description cs:0.95,0.0)},anchor=north},
        y label style={at={(axis description cs:-0.1,.5)},rotate=90,anchor=north},
        domain=0:10,
        samples=100,
        ticks=none,
        label style={font=\bfseries}, % Make all labels bold
        % Arrow style setting
        every axis plot post/.append style={line width=1.5pt, -{Latex[length=4mm,width=4mm]}}
    ]
    % Flat line
    \addplot[blue, domain=0:9] {0.03 + 0.25 * exp(-x/3)};
    \node at (axis cs: 6.1,0.16) [right, align=left, blue, font=\bfseries] {Importance of\\high early LR};
    % Exponential growth curve starting more flat then exploding
    \addplot[red] expression[domain=0:8.5,samples=50] {x < \growthxspikept ? (\growthyflatpt) : (\growthyflatpt + pow((x-\growthxspikept)/\growthdescale, \growthxspikept))};
    \node at (axis cs: 3.35,0.45) [right, align=left, red, font=\bfseries] {Importance of\\LR decay,\\weight decay};
    \end{axis}
\end{tikzpicture}
\mbox{}
\vspace{-6mm}
\mbox{}
\caption{\textbf{HP influence vs.\ TPP}: Higher TPP means higher
  gradient noise; LR decay \& weight decay settings thus increase in
  importance with TPP.\label{fig:cartoon}}
\vspace{-4mm}
\end{wrapfigure}
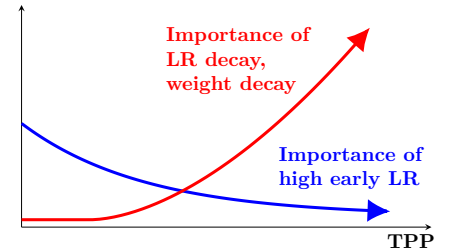

Per-step gradient variance is known to increase over the course of
training~\citep{mccandlish2018empirical}.
%showed that the ratio of
%gradient variance to gradient magnitude increases over the course of
%training, a ratio they equate to the \emph{critical batch size} (the
%batch size at which the benefits of parallelism rapidly decrease due
%to gradient redundancy).
%
Very recently, \citet{zhang2024how} showed that for an LLM of a fixed
size, training with larger datasets corresponds to larger
\emph{critical batch sizes}, which directly relates to larger
(aggregate) gradient variance via the gradient noise
scale~\citep{mccandlish2018empirical}.

In the EMA perspective, parameters $\theta_t$ are a combination of
prior weight updates (indexed by $i$).  The more variance at step $t$,
the more updates that should be combined in order to reduce it.
Here variance is reduced by combining updates \emph{across} steps,
rather than increasing batch size at a \emph{specific} step.
Now, with a constant LR, update coefficient $c_{t, i}$ decreases
exponentially in $(t - i)$: the further back the update from the
current step, the less it contributes.
But if LR $\eta_i$ \emph{decreases} over steps, each later coefficient
$c_{t, m}$ will be smaller due to a lower $\alpha_m = \eta_m\lambda$.
Yet since coefficients sum to one, \emph{earlier} coefficients will
correspondingly \emph{increase}: coefficients are basically flattened
and updates contribute more evenly across steps.
%
%corresponds to scaling
%\emph{up} all $c_{t, k}$ where $k < i$ (cf.~\cref{eqn:extended_ema}).
%This has the effect of flattening the coefficients, effectively
%averaging over more outputs.
%
The more we decay, the more updates we average over, and the more
variance is reduced.
%
%(see appendix \cref{fig:more_ema_lrs} for a visual
%contrast between $\constant$ and $\dtoz$ coefficients, on a log scale)

\paragraph{Bias reduction}
Higher decay is preferable to reducing \emph{peak} LR because we must
also minimize \emph{bias}, i.e., the contribution of the initial
random weights.
In the EMA perspective, the contribution of these weights to
$\theta_t$ is $c_{t,1} = \prod_{j=2}^{t} (1 - \alpha_j)$.  For a
decaying LR, and where $\alpha_j = \eta_j\lambda \ll 1$, coefficient
$c_{t,1}$ can be approximated (with equality when LR is constant,
\cref{sec:c_bias_derivation}):
\begin{equation}\label{eq:c_bias}
c_{t,1} \approx (1-\bar{\alpha})^{t-1}
\end{equation}
where $\bar{\alpha}$ is the average $\alpha_j$ over the schedule.
Exactly as in \cref{eqn:biasvar}, bias therefore decreases
exponentially in the \emph{absolute} number of steps, $t$, with a rate
of decrease depending on the LR\@.  Crucially, this means that as we
train for more total steps (i.e., a higher tokens-per-parameter; TPP),
there is a decrease in the \emph{fraction} of steps required for
$c_{t,1}$ to become negligible.  At higher TPP, bias reduction becomes
relatively less important than variance reduction
(\cref{fig:cartoon}).

\subsection{Experimental Hypotheses}

\hypothesis{As TPP increases, the relative benefit of $\dtoz$ over
  $\tenx$ decay will increase.}

This hypothesis follows from the premise that gradient variance plays
a larger role at higher TPP, and greater LR decay (as in $\dtoz$)
allows for more updates to be averaged, and thus greater variance
reduction.
A related hypothesis is that if we increase the batch size at each
step, gradient variance will decrease, and so the benefits of $\dtoz$
over $\tenx$ decay should \emph{diminish}.
Note our conceptual framework does not say precisely at which TPP or
batch size that $\dtoz$ will first prevail; here we will rely on our
empirical findings (\cref{sec:empirical}) to fill in the theoretical
gap.

\hypothesis{As TPP increases, the optimal peak LR will decrease, for
  all LR schedules.\label{hyp:optimal_lr}}

Tuning peak LR is about trading off movement from initial
conditions (requiring a high LR) and mitigating variance (requiring a
low LR).  As TPP increases, and bias reduction plays a smaller role,
optimal peak LR should decrease.  We hypothesize the decrease will
be greater with a $\constant$ schedule, as $\constant$ does not use
decay to balance the conflicting demands of bias and variance
(\cref{fig:bias_var_all}).
Moreover, optimal peak LR for other \emph{continuous} LR schedules,
like $\wsd$ and $\cyclic$, should also decrease with longer training
durations.  In this way, such ``schedule-free'' approaches are
\emph{not truly schedule-free}.  This dependence is also obvious when
plotting update coefficients for these schedules (appendix
\cref{fig:inf_ema_lrs}): the higher the LR, the more emphasis on
recent updates.

\hypothesis{$\linear$ $\dtoz$ will improve over $\cosine$ $\dtoz$.}

While LR decay allows averaging over more weight updates, if the LR
decreases too quickly, the final weight updates may not contribute to
the EMA\@.  From a loss surface perspective, as the LR approaches
zero, we take vanishingly small steps.  Since $\cosine$ reaches
smaller steps faster than $\linear$ (\cref{fig:background_ema}, left),
$\cosine$ will make less progress toward the optimum loss.
From the EMA perspective, this is equivalent to $\cosine$ having
smaller $c_{t,i}$ coefficients as $i$ approaches $t$
(\cref{fig:background_ema}, right).
Note this problem is unique to $\cosine$ $\dtoz$ and will not affect,
e.g., $\cosine$ $\tenx$ decay.

\hypothesis{A high LR (not weight decay) reduces bias and achieves optimal loss.\label{hyp:wdbias}}

Note that weight updates $x_t$ have a coefficient of
$\nicefrac{1}{\lambda}$ in \cref{eqn:adamwema}.  So, while $\eta_t$
and $\lambda$ contribute equally to $\alpha_j$, increasing $\lambda$
to reduce bias is counterproductive as weight updates will be scaled
down proportionally, reducing movement from initial conditions.
%This observation will be crucial for interpreting our experimental
%findings where we systematically vary LR and weight decay (e.g.,
%\cref{fig:maxlr_wd:617M}).
%
However, if LR is in a high enough range so that bias plays a minimal
factor, both LR and WD should equally affect variance reduction.

%Since weight decay does not impact bias reduction, then at high TPP,
%it should be more effective to reduce variance by lowering WD than by
%reducing the LR\@.
%%
%This stands in contrast to very recent work that recommends purely
%decreases in LR for high-TPP training~\citep{bjorck2024scaling}.

\section{Empirical Analysis of Decay-to-Zero}\label{sec:empirical}
\subsection{Experimental setup}

% Add this back in later: LM~\citep{dey2023btlm3b8k}
Experiments use a GPT-like LLM~\citep{radford2019gpt2}, with ALiBi
embeddings~\citep{press2022alibi} and SwiGLU~\citep{shazeer2020glu}.
Main paper models are trained on
SlimPajama~\citep{cerebras2023slimpajama} and evaluated over 1.1B
held-out tokens (regardless of training TPP).
Unless otherwise indicated, $\lambda$=$0.1$ is used.
Training runs use the same random seed, so all decay functions
($\linear$, $\cosine$, etc.) and ratios ($\constant$, $\tenx$,
$\dtoz$) have identical warmups, but note results are very consistent
across seeds at this scale (appendix \cref{fig:seeds}).
By default we use $\mup$ (standard parameterization results are in
\cref{subsec:sp,subsec:nanogpt}).  $\mup$ HPs are derived from a
smaller proxy model tuned using a $\linear$-$\tenx$ schedule.
%
%(results for $\dtoz$ could conceivably
%improve if we tuned for it instead).
%
Since we hypothesized that different LR schedules may enjoy their
optimal peak LR at different values, our experiments compare schedules
when each is tuned to its optimal peak LR\@; we sweep $\hateta$ by
factors of 2$\times$ around the $\mup$ proxy-tuned $\hateta = \maxlr$.
\cref{sec:experimental_details} provides further details on the model
architecture (\cref{tab:model_info}), dataset sizes
(\cref{tab:train_steps}), and compute resources.

\subsection{Results}

\begin{table}
  \centering
  \caption{\textbf{Comparison of schedules}: validation loss for 610M
    $\mup$ models, 20~TPP\@.  $\hateta$=$\maxlr$ is the $\mup$
    proxy-tuned peak (base) LR, $\eta$ is the LR after $\mup$
    adjustment ($\rho$=$0.125$, see \cref{subsec:mup}). $\dtoz$ is
    superior to $\tenx$ decay across most LRs. $\linear$-$\dtoz$
    slightly outperforms $\cosine$-$\dtoz$ in all
    cases.\label{tab:sched_compare}}
  \begin{tabular}{@{}cccccccc@{}}
    \toprule
    \multirow{2}{*}{$\hateta$} & \multirow{2}{*}{$\eta$} &\multirow{2}{*}{$\invsqrt$} & \multirow{2}{*}{$\constant$} & $\cosine$ & $\linear$ & $\cosine$ & $\linear$ \\
                               &                         &                            &                              &   $\tenx$ & $\tenx$   & $\dtoz$   & $\dtoz$   \\ \midrule
    $6.5$e-$02$ & $8.1$e-$03$ & 2.789      & 3.035       & \texttt{NaN}               & 2.667             & 2.611             & \textbf{2.605}    \\
    $3.2$e-$02$ & $4.0$e-$03$  & 2.710      & 2.850       & 2.604             & 2.606             & 2.574             & \textbf{2.571}    \\
    $\maxlr$* & $2.0$e-$03$ & 2.671      & 2.768       & 2.590             & 2.591             & 2.578             & \textbf{2.573}    \\
    $8.1$e-$03$ & $1.0$e-$03$ & 2.665      & 2.722       & 2.598             & 2.600             & 2.595             & \textbf{2.590}    \\
    $4.0$e-$03$ & $5.1$e-$04$ & 2.691      & 2.711       & 2.634             & 2.635             & 2.637             & \textbf{2.633}    \\
    $2.0$e-$03$ & $2.5$e-$04$ & 2.762      & 2.739       & \textbf{2.707}    & 2.710             & 2.717             & 2.714             \\ \bottomrule
  \end{tabular}
\end{table}

\finding{$\linear$-$\dtoz$ is the optimal LR schedule across virtually
  all peak learning rates.}

For 610M-parameter $\mup$ models trained to compute-optimal 20~TPP,
the optimal $\linear$-$\dtoz$ setting achieves 0.77\% lower loss than
the optimal $\linear$-$\tenx$ setting (\cref{tab:sched_compare}).
At smaller, suboptimal peak LRs, $\tenx$ can be better than $\dtoz$.
In appendix \cref{fig:maxlr_sp.curves}, we demonstrate very
similar results using the standard parameterization.
Regarding the decay function, gains from $\linear$-$\dtoz$ over
$\cosine$-$\dtoz$ are small, but perfectly consistent across peak LRs,
exactly in line with recent
work~\citep{defazio2023when,lingle2024large}.
Interestingly, $\cosine$-$\tenx$ is consistently slightly better than
$\linear$-$\tenx$, showing that $\linear$ itself is not always best,
rather $\linear$ \emph{plus} $\dtoz$ is needed.
Lacking a cooldown phase, inverse square root ($\invsqrt$) and
$\constant$ do not perform as well.
%
%\citep{lingle2024large} - if you look through his findings, you also
%see that $\linear$ slightly beats $\cosine$ (across all LRs), except at the
%very smallest size.  This wasn't the point of that work, but note it!

Appendix \cref{fig:maxlr_tpp:models} compares schedules as we sweep
peak LR for 111M, 610M, and 1.7B models.  For these and subsequent
experiments, we use a $\linear$ decay.

\begin{figure}
  \vspace{-2mm}
  \centering
  \scalebox{0.85}{
    \includegraphics[width=\textwidth]{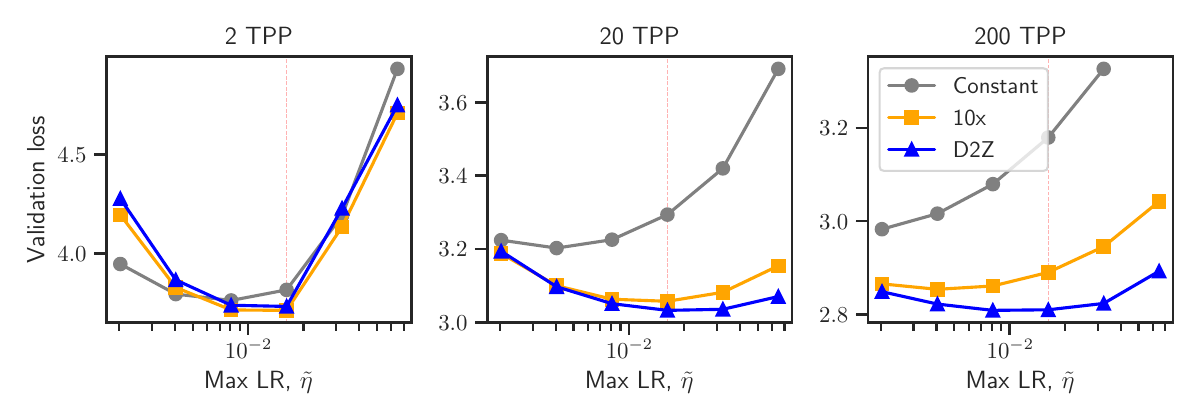}
  }
  \mbox{}
  \vspace{-4mm}
  \mbox{}
  \caption{\textbf{Comparing decay schedules across TPP} (111M
    scale): As TPP increases, $\linear$-$\dtoz$ outperforms $\tenx$,
    especially at the proxy-tuned peak LR (red lines).
    %610M results are similar (appendix \cref{fig:maxlr_tpp:610M}).
    \label{fig:maxlr_tpp:111M}}
\end{figure}

\finding{As TPP increases, the relative improvement of $\dtoz$ over
  $\tenx$ also increases.}

For 111M models at only 2~TPP, $\dtoz$ performs \emph{worse} than
$\tenx$ across all peak LRs (\cref{fig:maxlr_tpp:111M}, left).  610M
results are similar (appendix \cref{fig:maxlr_tpp:610M}).  However, as
TPP increases, $\dtoz$ begins to outperform $\tenx$, exceeding the
best setting of $\tenx$ by 1.6\% at 200~TPP, and performing 2.8\%
better at the proxy-tuned $\hateta$=$\maxlr$ (marked in plots with a
red vertical line).
While we hypothesized $\dtoz$ would surpass $\tenx$ at \emph{some}
TPP, our key finding is that, at both compute-efficient (20~TPP) and
over-trained sizes, decaying to \emph{zero} is optimal.
Even at 2~TPP, some LR decay is beneficial ($\tenx$ improves over
$\constant$), but too much evidently hampers bias reduction.
As TPP increases, optimal LRs for $\tenx$ shift lower as variance
reduction gains in importance, but now it is the lower \emph{peak} LRs
that hamper bias reduction, and $\tenx$ begins to lag $\dtoz$.
By 20+~TPP, $\dtoz$ is consistently superior.
%
%Discuss this a bit:
%
%With $\constant$, the shifting emphasis as TPP increases can only be
%satisfied by lowering the fixed LR (explaining the shift in optimal
%LR, \cref{fig:maxlr_tpp:111M,fig:maxlr_tpp:610M}).

\cref{fig:more_tpp} plots validation loss of $\tenx$ and $\dtoz$ at
different TPP settings for different model sizes (all models are
trained using the proxy-tuned peak LR\@).
For the 610M model, $\dtoz$ is also initially worse than $\tenx$, but
begins to surpass it around 4~TPP, and by 200~TPP is 2.6\% better.  As
with training loss (\cref{fig:train_loss}), an 80~TPP $\dtoz$
model can surpass a 200~TPP $\tenx$ model in validation loss.
\textbf{Importantly, these trends also hold for \emph{downstream}
  evaluation of the models} (appendix \cref{tab:downstream1}).

\cref{fig:more_tpp} also shows the advantage of using $\dtoz$ with
\emph{1.7B models trained to 80 TPP}\@; $\dtoz$ achieves roughly
3.05\% lower loss than $\tenx$ at the proxy-tuned $\hateta$.  At this
scale, such gains provide real-world impact: we over-train models of
this size for use as proposal models in speculative
decoding~\citep{leviathan2023fast} and smaller models like
Gemma-2B~\citep{mesnard2024gemma} and Phi-1.5~\citep{li2023textbooks}
are used in many production settings.
%
%So, finding the right recipes
%for over-training models at this scale is an important research
%direction and a valuable supplementary contribution of this paper.

\begin{figure}
\noindent % Ensures the minipages align with the left margin
\begin{minipage}[b]{0.39\textwidth}
  \centering
  \input{fig_more_tpp.tex}
\end{minipage}%
\hfill % Use to fill space between the minipages
\begin{minipage}[b]{0.57\textwidth}
  \centering
  \input{fig_maxlr_isotppbatches.tex}
\end{minipage}
\end{figure}

\finding{Compared to $\constant$ and $\tenx$, optimal peak LRs are much
  more stable with $\dtoz$.\label{find:optimal_lr}}

\cref{fig:maxlr_tpp:111M} reveals different levels of hyperparameter
sensitivity when increasing TPP\@: optimal LR shifts substantially
lower for $\constant$, somewhat lower for $\tenx$, and hardly at all
for $\dtoz$.  The same trend holds at the 610M scale (appendix
\cref{fig:maxlr_tpp:610M}).
Moreover, with $\dtoz$, loss is less sensitive to a suboptimal LR
(bowls are flatter).
$\dtoz$ loss is also more stable (and lower) as we vary both
\emph{weight decay} (appendix
\cref{fig:maxlr_wdgroups:610M,fig:maxlr_wdgroups:111M}) and
\emph{batch size}, $B$.
For example, the optimal LR already shifts significantly for $\tenx$
models at B=126, while $\dtoz$ optima only begin to shift at B=63
(\cref{fig:maxlr_isotppbatches}).  The superiority of $\dtoz$ over
$\tenx$ across $B$ is more clearly observed in appendix
\cref{fig:maxlr_isotppbatches_sep_batches}.

Crucially, if we keep the number of optimization steps constant
(11752), but only vary $B$ (effectively increasing TPP proportional to
$B$), we find the optimal learning rate does not \emph{decrease}
(i.e., with TPP) but rather \emph{increases} (appendix
\cref{fig:maxlr_isostepbatches}).  Clearly, LR decay is primarily
beneficial as a noise-reduction mechanism: with larger batches, we
have less gradient noise, and can afford a larger LR\@.

\finding{$\linear$-$\dtoz$ works better than $\wsd$ and $\cyclic$
  (continuous) LR schedules.}

\begin{figure}
  \noindent
\begin{minipage}[b]{0.40\textwidth}
  \centering
  \input{fig_maxlr_infinite.tex}
\end{minipage}%
\hfill
\begin{minipage}[b]{0.58\textwidth}
  \centering
  \input{fig_trainevals.tex}
\end{minipage}
\end{figure}

In \cref{fig:maxLR_infinites}, we compare $\dtoz$ to $\tenx$, and to
two approaches designed for continuous pre-training: $\cyclic$, which
cycles LR up and down, and $\wsd$ (\cref{subsec:lr_schedules}).  For
$\wsd$, we simulate a model being retrieved after 80~TPP, and cool the
LR to zero for the final 22.5\% of steps (around the proportion
recommended by \citet{hagele2024scaling}, and equal to the cooldown
duration in our 20~TPP models).  Appendix \cref{fig:inf_ema_lrs}
provides full LR curves and update coefficients for all models in
\cref{fig:maxLR_infinites}.

At its optimal LR, $\wsd$ works better than $\tenx$, confirming
results in \citet{hagele2024scaling}.  However, note the optimal peak
LR shifts lower for both $\tenx$ and $\wsd$ at this TPP\@.
$\linear$-$\dtoz$ remains best here, around 0.84\% better than the
optimal $\wsd$.  Given the diminishing returns of high-TPP training
(\cref{fig:more_tpp}), $\wsd$ would require significantly more
training FLOPs to reach the level of $\dtoz$.

\finding{$\constant$ and $\tenx$, but not $\dtoz$, strongly overfit to
  the end of the training data.}

\cref{fig:trainevals} shows the loss of trained models (with frozen
weights) on the exactly-same ordered batches used during pre-training.
As predicted by the extended EMA perspective (\cref{sec:emas}), the
higher the LR, the more models fit to late-stage training batches.  It
is striking that both $\constant$ and $\tenx$, but not $\dtoz$,
\emph{overfit} to the very final portion.  Extra adaptation of
generative models to recent training sequences has long been
observed~\citep{graves2013generating}, but to our knowledge this is
the first evidence $\dtoz$ may help mitigate these effects.  Since
$\dtoz$ demonstrates the lowest loss on batches \emph{slightly before}
the final training phase, placing the highest-quality and most-recent
data in the \emph{very} final phase, while using $\dtoz$ (e.g., as
in~\citet{dubey2024llama}), may be suboptimal.
These findings also contradict \citet{biderman2023pythia}, who found
training order had little impact on memorization.

\section{Discussion}\label{sec:discussion}
\paragraph{The confounding role of LR schedule}

We have shown that the more constant the LR schedule, the more
sensitive the optimal peak $\eta$.
%(and the more the model overfits the end of the training data.
%provides empirical evidence for this perspective; the highest peak
%LRs achieve lowest loss on the final (re-visited) training batches.
These findings help explain LR sensitivity in prior work.
For example, \citet{shen2024power} were puzzled by their observation
that ``although the $\wsd$ scheduler could, in theory, continue in the
stable phase forever \ldots the optimal LRs are different for different
amounts of training tokens.''
As noted under \cref{hyp:optimal_lr}, LR schedules with long constant
periods only appear ``schedule-free'' in the primal; from the dual
(EMA) perspective, the higher the LR, the more emphasis is placed on
recent updates (see appendix
\cref{fig:more_ema_lrs:constant} for $\constant$, 
\cref{fig:inf_ema_lrs:wsd} for $\wsd$).
\citet{filatov2024time} and \citet[Figure~19]{yang2022mup} also
observed major decreases in optimal LR with higher TPP; these results
were both obtained with a constant LR\@.
In contrast, \citet{bjorck2024scaling} observed less significant
decrease of optimal $\eta$ with TPP, but used $\tenx$ decay.

LR schedule may also affect the optimal $\eta$ as batch size varies
--- particularly when using $\mup$.
Some prior $\mup$
studies~\citep{yang2022mup,noci2024learning,shen2024power} observed
linear scaling of optimal LR with batch size, i.e., the
so-called \emph{linear scaling
rule}~\citep{krizhevsky2014one,chen2016revisiting,smith2018dont}.
Others~\citep{lingle2024large} have observed square-root scaling,
resonating with other prior
studies~\citep{hoffer2017train,you2019large,malladi2022sdes}.
% You should note that the original $\mutransfer$
%paper~\citet{yang2022mup} looked at transferability at iso-Step cases
%(footnote 20), whereas \citet{noci2024learning} tried iso-TPP, but I
%no longer think this is a key issue.
This discrepancy can be explained by noting the linear scaling results
were all found with a $\constant$ or $\wsd$ LR decay, while
square-root was observed with $\linear$ $\dtoz$, again underscoring
the greater stability of $\dtoz$.

\paragraph{The confounding role of training duration}\label{subsec:confounding}

While LR schedules have confounded studies varying TPP, TPP has
analogously confounded studies evaluating LR schedules.
Recall \citet{kaplan2020scaling} also saw a benefit from $\dtoz$.  In
contrast to Chinchilla scaling~\citep{hoffmann2022empirical}, in
the \citeauthor{kaplan2020scaling} perspective, small models should be
trained to high TPP (while larger models should be trained less).  It
is therefore not surprising \citeauthor{kaplan2020scaling} saw
benefits testing $\dtoz$ with small models; as we have shown, $\dtoz$
is especially effective at high TPP\@.
In contrast, in Figure~4 of \citet{yang2022mup}, $\linear$
is \emph{worst} of all schedules, and the gap between it and
$\constant$ and $\invsqrt$ grows with model width.  But here, since
training data is small --- and \emph{fixed}, then TPP
\emph{decreases} as width increases.  Thus training is in a phase
where bias reduction is paramount and $\dtoz$ is not effective.
Notably, in their LLM training experiments at higher TPP,
\citeauthor{yang2022mup}\ do report linear $\dtoz$ to work best.
%
%This is the right time to revisit these:
%\citet{shallue2019measuring}
%compared many different LR schedules (cosine, linear, piecewise,
%polynomial, etc.) on ResNETs.  They ultimately recommended linear
%decay as a simple approach that works as well as the best alternative,
%and requiring fewer hyperparameters to tune.
%
%\citet{hoffmann2022empirical} adopt cosine decay, and report that
%``the difference between decaying by 10 and decaying to 0.0 ... is
%small, though decaying by a factor of 10 to be slightly more
%performant,'' while ``decaying by less (5) is clearly worse.''

With this context, we speculate on why $\dtoz$ has not been adopted
more widely in LLM pre-training:
\begin{itemize}[leftmargin=*]
  \item First, it is common to evaluate hyperparameters on smaller
    training runs; unfortunately, with limited training data (low
    TPP), $\dtoz$ misleadingly under-performs.
% , or partway through training (e.g., \citet{almazrouei2023falcon}
%choose their peak LR based on loss rankings at the end of the warmup
%period).  ``Loss rankings at the end of learning rate warm-up broadly
%reflects rankings at the end of training, enabling us to search for
%optimal learning rates efficiently''~ – i.e., evaluate them only at
%end of warm-up phase.
\item Secondly, coupling between LR schedule and optimal peak LR is
  problematic: with $\tenx$ decay, we may find a lower peak LR is
  optimal; if we then test $\dtoz$ with the same LR, it may be
  suboptimal for $\dtoz$.  Not seeing a benefit at this LR, we may
  conclude $\dtoz$ is inferior in general.
\item Finally, poorly controlled training dynamics may prevent
  $\dtoz$ models from training with their (higher) optimal LR: When we
  initially compared $\dtoz$ and $\tenx$ using NanoGPT
  (\cref{subsec:nanogpt}), $\tenx$ performed better at the default
  LR\@.  Raising the LR resulted in training divergence.  Only after
  switching the numerical precision from \verb|bfloat16|
  to \verb|float32| could $\dtoz$ succeed --- and the model reach
  optimal loss.
\end{itemize}

%
%Things that enable stability, see~\citet{wortsman2023small} -- e.g.,
%longer warmup.  QK-layernorm, z-loss regularization, decoupled weight
%decay, width vs. depth, mup.
%
%``We briefly revisited the decay
%schedule (keeping the decay style constant to a cosine decay),
%ablating decaying to 0 vs. 10\% of the learning rate. Decaying to 0
%yielded slightly better downstream scores, so we chose this value.''
%
%However, it seems recently, the merits of $\dtoz$ are becoming more
%widely appreciated, and being used more in SOTA frontier
%models~\citet{alephalpha2024introducing,dubey2024llama}.
%%
%We hope it may also become the reference for schedules for continuous
%pre-training~\citep{hagele2024scaling}.

%Basically we must acknowledge that linear D2Z is starting to catch on.
%LLaMA3, Alepha Alpha.  WSD usually goes to zero.  But the baselines
%should also D2Z.  Plus growing theory supporting D2Z.

%In fact, I now see it as almost a diagnostic whether D2Z helps.  If it
%doesn't, your peak LR is likely lower than optimal, and you need to
%work on stabilizing training so you can push it higher.  But when you
%do, D2Z will be there, waiting for you.

\paragraph{The special benefit of $\dtoz$}

%Thus far, the dual view of the LR schedule has provided some theory
%explaining the slight, consistent benefits of $\linear$ over
%$\cosine$ (\cref{fig:background_ema},
%\cref{tab:sched_compare})... Going back to its origination
%in~\citet{loshchilov2016sgdr}, we are not aware of any theory
%supporting the use of $\cosine$.  You can't just drop the LR at the
%end (Step decay).  It explains the dependence of $\wsd$ and
%$\constant$ on the peak LR.  The fact the different curves hit their
%minimum at the same $\alpha$, suggests this is an important quantity.
%But the fact they are separated based on their $\hateta$ is also
%crucial.  And the fact there's a cooldown at the end, that's not
%really accounted for.
%People are putting their high-quality and recent data at the end of
%training (LLaMA3).  That makes sense then.  Actually, it was observed
%that the final data you train on influences the character of generated
%text as far back as Alex Graves's work.

We have shown that a low $\alpha$ later in training can expand the
timescale over which weight updates are combined, reducing noise in a
similar way to increasing batch size.
However, there is apparently a separate, independent benefit from a
vanishing LR\@.
% but only when the models are sufficiently far from the initial
% conditions
Indeed, looking at appendix \cref{fig:inf_ema_lrs} for the 80~TPP
comparison to continuous schedules, we see $\tenx$ coefficients are
quite similar to the $\dtoz$ curve, apart from the final drop.
Moreover, they are flatter than $\wsd$ EMA coefficients for the
same peak LR, suggesting better integration of prior updates.  Yet
$\wsd$ performs \emph{better} than $\tenx$ at all LR settings
(\cref{fig:maxLR_infinites}).
In contrast, at 2~TPP (\cref{fig:maxlr_tpp:111M}, left), $\tenx$
performs better than $\dtoz$ at every LR setting.
Prior work has shown large LRs allow exploration of the loss surface
at a height ``above the valley floor''~\citep{xing2018walk}, while LR
cooldown phases descend into a local
minimum~\citep{hu2024minicpm,hagele2024scaling}.
It appears descending into these minima is beneficial only after
sufficient exploration of the loss surface.

\section{Limitations and Further Experiments}

While our findings strongly support linear $\dtoz$ being optimal in
our specific context, there are some limitations to keep in mind.
First, LR schedules like $\dtoz$ require prior knowledge of the total
number of training steps.  But it is worth reiterating that even
nominally schedule-free approaches such as $\constant$ also require
this knowledge in order to optimally set the LR value.  In contrast,
the extended EMA perspective of LR schedules enables derivation of a
truly schedule-free schedule, which we introduce
in \cref{subsec:rational}.
Second, our focus in this paper was specifically LLM training at
compute-optimal and overtrained dataset sizes.  For ML problems with
limited access to training data, $\dtoz$ is likely not the best
strategy.
Third, our work focuses on AdamW (the standard optimizer in LLM
pre-training).
While the extended EMA perspective of LR schedules will likely apply
to other optimizers that use decoupled weight decay (as noted
by~\citet{wang2024how}), it may not apply to approximate second order
methods, such as Shampoo~\citep{gupta2018shampoo}.
Finally, for LLMs with unstable training dynamics that cannot tolerate
high LRs, $\dtoz$ may not be beneficial.  We experienced this
first-hand when we initially trained NanoGPT~(\cref{subsec:nanogpt}).

However, we also note the remarkable consistency of $\dtoz$'s success.
While results in the main paper used the SlimPajama dataset,
consistent results were found with OpenWebText (\cref{subsec:nanogpt})
and a multilingual dataset (with a larger vocabulary)
(\cref{subsec:scaling}).
%Understanding if there is a relationship between dataset quality and
%the optimal decay schedule is an avenue for future research.  with
%downstream evaluations (\cref{subsec:downstream}),
We also saw consistent findings with different parameterizations
(\cref{subsec:sp,subsec:nanogpt}), architectures
(\cref{subsec:scaling}), weight sparsity settings
(\cref{subsec:sparsity}), and training frameworks
(\cref{subsec:nanogpt}).
\cref{subsec:scaling} describes scaling laws fit to models trained
with $\dtoz$ and $\tenx$, at model sizes up to 2.75B; results indicate
a growing performance gap as scale increases.
Finally, \cref{subsec:wd} demonstrates that the benefits
of \emph{weight decay} are observed primarily when using LR $\dtoz$,
where raising weight decay can fine-tune the EMA update coefficients
without affecting training stability.

\section{Conclusion}\label{sec:conclusion}
Linear decay-to-zero is the optimal LR schedule for LLM training using
AdamW\@.
To be clear, less decay is beneficial for low tokens-per-parameter
training, but there is no practical reason to perform such training
with LLMs, since the same FLOPs could be used to train a smaller
model, over more tokens, to a lower loss -- using $\dtoz$.
The superiority of $\dtoz$ was validated across a range of
experimental conditions.  Results suggest its relative benefit will
increase as models increase in scale.  Moreover, when using $\dtoz$
and $\mup$, the optimal peak LR is less sensitive to changes in weight
decay, dataset size, and batch size, i.e., there is better
hyperparameter \emph{transfer}.

%Varying the decay schedule has proven to be a useful tool for
%stress-testing LLMs and developing insights into training.  Here,
%
Our analysis was aided by our interpretation of AdamW's output as a
convex combination of prior weight updates.
$\dtoz$ overfits less the final training sequences, and is especially
beneficial when gradient noise dominates training.
As we enter a phase of applied ML where inference efficiency is a
primary concern, there is strong motivation to study high-TPP
training, where gradient noise is the bottleneck.
While our results indicate that $\dtoz$ is a key component of the
solution here, further investigation is required, including into how
and when to adjust hyperparameters such as weight decay, batch size,
and learning rate, in the high-TPP context.

%\subsubsection*{Author Contributions}
%If you'd like to, you may include  a section for author contributions as is done
%in many journals. This is optional and at the discretion of the authors.

%\subsubsection*{Acknowledgments}
%Use unnumbered third level headings for the acknowledgments. All
%acknowledgments, including those to funding agencies, go at the end of the paper.

\bibliography{decay2zero}
\bibliographystyle{cereb}

\newpage
\appendix
\section{Experimental Details}\label{sec:experimental_details}

\begin{table}
  \centering
  \caption{\textbf{Model architecture and batch sizes for main experiments}\label{tab:model_info}}
\begin{tabular}{@{}ccccccc@{}}
\toprule
Model & vocab. size & $d_{model}$ & $n_{layers}$ & $d_{head}$ & $\dffn$ & batch size \\ \midrule
111M  & 50257       & 768         & 10           & 64         & 2048    & 192        \\
610M  & 50257       & 2048        & 10           & 64         & 5461    & 504        \\
1.7B  & 50257       & 2048        & 32           & 64         & 5461    & 504        \\ \bottomrule
\end{tabular}
\end{table}

\begin{table}
  \centering
  \caption{\textbf{Training steps for main experiments}\label{tab:train_steps}}
\begin{tabular}{@{}ccccc@{}}
\toprule
Model & TPP & Warmup & Steps  & Tokens \\ \midrule
111M  & 2   & 56     & 557    & 219M   \\
111M  & 20  & 556    & 5568   & 2.19B  \\
111M  & 200 & 5560   & 55680  & 21.9B  \\
610M  & 2   & 118    & 1176   & 1.21B  \\
610M  & 20  & 1175   & 11752  & 12.1B  \\
610M  & 80  & 4700   & 47008  & 48.5B   \\
610M  & 200 & 11750  & 117520 & 121B   \\
1.7B  & 20  & 3322   & 33220  & 34.3B  \\
1.7B  & 80  & 13288  & 132880 & 137B  \\ \bottomrule
\end{tabular}
\end{table}

\cref{tab:model_info} provides details on model architecture and
hyperparameters for the main experiments (i.e., results presented in
the main paper).  \cref{tab:train_steps} provides information on
the training steps.  All the models in our main experiments were
trained on the SlimPajama dataset~\citep{cerebras2023slimpajama}, a
cleaned and deduplicated version of the RedPajama dataset.  We use the
GPT-2~\citep{radford2019gpt2} vocabulary of size 50257, and a context
length of 2048 tokens.
Unless otherwise noted, the weight decay value, $\lambda$, is by
default set to 0.1, following standard practice in LLM
pre-training~\citep{brown2020language,hoffmann2022empirical,almazrouei2023falcon,alephalpha2024introducing}.
Also following standard practice, we do not apply weight decay or bias
to LayerNorm layers.
Validation loss is always computed over 1.1B held-out tokens,
regardless of training TPP\@.
By default we parameterize with $\mup$ (further details below).

For a given TPP, all models have the exact same warmup phase: a linear
warmup of the LR from 0 to the peak value.
In all our runs, warmup was 10\% of the total steps.
LR warmup is standard practice in LLM pre-training.\footnote{While
prior work has suggested LR warmup is less valuable in modern Pre-LN
Transformers~\citep{xiong2020layer}, various other studies have shown
warmup leads to lower
loss~\citep{goyal2017imagenet1hour,liu2019variance,tissue2024scaling,kosson2024analyzing},
and may reduce sensitivity to peak LR~\citep{wortsman2023small}.  In
light of the similar benefits of $\dtoz$, it would be interesting to
investigate the value of warmup for models that are specifically
trained using $\dtoz$.}

All models in the main experiments were trained on a Cerebras CS-3
system.  610M-parameter, 20~TPP models take roughly 6 hours each to
train on a single CS-3.
If a training run did not complete due to numerical instabilities, the
values are left off our plots or marked as \texttt{NaN} in
\cref{tab:sched_compare}.
%
%Give a table with our $\mup$ changes.

% Tissue: Our pilot experiment (refer to Appendix A) shows that warmup
% indeed significantly accelerates convergence, a finding also noted
% by Liu et al. (2020); Kosson et al. (2024).
% Li: "The learning rate warmup heuristic achieves remarkable success in stabilizing training"
% Kosson: "This aids training"

% Ibrahim: Goyal et al. (2018) found that a gradual warm-up of LR
%early on in training can help overcome optimization challenges,
%particularly with large mini-batch sizes. Additionally, Popel & Bojar
%(2018) emphasized the importance of a warm-up stage when training
%Post-LN Transformers.  On the other hand, Xiong et al. (2020)
%discovered that Pre-LN Transformers are more stable and may not
%require a warm-up stage.
%Longer warmup reduces LR sensitivity!~\citep{wortsman2023small} -- why
%might that be?  I mean, it does move you away from the initial
%conditions... 
%
%Might be worth perusing: Justin Gilmer, Behrooz Ghorbani, Ankush Garg,
%Sneha Kudugunta, Behnam Neyshabur, David Cardoze, George Dahl, Zachary
%Nado, and Orhan Firat. A loss curvature perspective on training
%instability in deep learning. arXiv preprint arXiv:2110.04369,
%2021. -- something about warmups helping.  But warmup isn't the point
%of your work.

\paragraph{Proxy model hyperparameter tuning}\label{sec:proxy_tuning}

\begin{table}
    \centering
    \caption{\textbf{Tuned hyperparameters for $\mup$ proxy model}\label{tab:mup_hps}}
    \begin{tabular}{cc}
         \toprule
         $\sigma_{W,\text{base}}$& $8.67$e-$02$ \\
         $\hateta$& $\maxlrdetail$\\
         $\alpha_{\text{input}}$& $9.17$\\
         $\alpha_{\text{output}}$& $1.095$\\
         \bottomrule
    \end{tabular}
\end{table}

To find the optimal $\mup$ hyperparameters (HPs), we trained a
39M-parameter proxy model using a width $d_{\text{model}}$=$d_p$ of
256, with 24 layers and a head size of 64.  We trained this proxy
model on 800M tokens with a batch size of 256 and context length 2048,
using $\tenx$ decay.  We randomly sampled 350 configurations of base
learning rates, base initialization standard deviation, and embedding
and output logits scaling factors, and used the top-performing values
as our tuned HPs (\cref{tab:mup_hps}).

\section{Derivations}\label{sec:derivations}

\subsection{Derivation of \cref{eq:c_bias}}\label{sec:c_bias_derivation}

The initial coefficient is $c_{t,1} = \prod_{j=2}^{t} (1 - \alpha_j)$.
Clearly, if $\alpha_j = \alpha$ is constant, we have $c_{t,1} = (1 - \alpha)^{t-1}$.
Otherwise, given $\alpha_j$ is small, we can use a first-order Taylor expansion,
$e^{-\alpha_j} \approx 1 - \alpha_j$, and therefore:
\begin{align*}
  c_{t,1} &\approx \prod_{j=2}^{t} e^{-\alpha_j} \\
  &= e^{\sum_{j=2}^{t} -\alpha_j} \\
  &= e^{- \sum_{j=2}^{t} \alpha_j}. \\
\end{align*}
Assuming we only know the average value, $\baralpha = \frac{1}{t-1} \sum_{j=2}^{t} \alpha_j$,
  we have:
\begin{align*}
  c_{t,1} &\approx e^{-\baralpha(t-1)} \\
          &= (e^{-\baralpha})^{t-1}. \\
\end{align*}
Given $\baralpha$ is also small, we reverse the earlier Taylor
expansion, but now $e^{-\baralpha} \approx 1 - \baralpha$, and:
\begin{align*}
  c_{t,1} &\approx (1 - \baralpha)^{t-1}. \\
\end{align*}

That is, with a small time-varying smoothing parameter, the initial
coefficient in an EMA, $c_{t,1}$, also decreases exponentially with
the number of steps.

\section{Additional experimental results}

In this section, we include some additional results to support the
findings in the main paper.  All validation losses reported in this
section are from models trained with $\linear$ decay.

\subsection{Downstream evaluations}\label{subsec:downstream}

\begin{table}
  \centering
  \caption{\textbf{Downstream evaluations}: for 610M-parameter models
    corresponding to \cref{fig:train_loss}.  A model trained for 80
    tokens-per-parameter with linear $\dtoz$ has equivalent downstream
    loss to the same model trained for 200~TPP with $\tenx$ decay.
    \label{tab:downstream1}}
    \fontsize{7.8pt}{9pt}\selectfont
    \setlength{\tabcolsep}{2pt}
\begin{tabular}{@{}lccccccccccccc@{}}
\toprule
\textbf{}  & \textbf{\begin{tabular}[c]{@{}c@{}}MMLU\\ (Avg.)\end{tabular}} & \multicolumn{8}{c}{\textbf{Commonsense Reasoning}}                                                                                                                                                                                                                                                                                                                                                     & \textbf{\begin{tabular}[c]{@{}c@{}}Reading\\ Comp.\end{tabular}} & \multicolumn{2}{c}{\textbf{\begin{tabular}[c]{@{}c@{}}Truthfulness\\ \& Bias\end{tabular}}}                                          & \textbf{\begin{tabular}[c]{@{}c@{}}\cellcolor{gray!30}Down-\\\cellcolor{gray!30}  stream\\\cellcolor{gray!30} (Avg.)\end{tabular}} \\ \midrule
           & \textbf{}                                                      & \textbf{\begin{tabular}[c]{@{}c@{}}Wino-\\ grande\end{tabular}} & \textbf{\begin{tabular}[c]{@{}c@{}}Hella-\\ swag\end{tabular}} & \textbf{\begin{tabular}[c]{@{}c@{}}Open-\\ Book\\ QA\end{tabular}} & \textbf{\begin{tabular}[c]{@{}c@{}}Lamb-\\ ada\\ OpenAI\end{tabular}} & \textbf{\begin{tabular}[c]{@{}c@{}}Lamb-\\ ada\\ Stand.\end{tabular}} & \textbf{SIQA} & \textbf{PIQA} & \textbf{Arc-e} & \textbf{RACE}                                                    & \textbf{\begin{tabular}[c]{@{}c@{}}Truth-\\ ful\\ QA\end{tabular}} & \textbf{\begin{tabular}[c]{@{}c@{}}CrowS-\\ Pairs\end{tabular}} & \cellcolor{gray!30}                                                                           \\
$\tenx$@80TPP  & 23.6\%                                                         & 52.2\%                                                          & 43.9\%                                                         & 31.4\%                                                             & 46.7\%                                                                & 36.7\%                                                                & 32.8\%        & 68.7\%        & 47.6\%         & 32.3\%                                                           & 39.8\%                                                             & 60.7\%                                                          & \cellcolor{gray!30}43.05\%                                                                   \\
$\dtoz$@80TPP  & 23.5\%                                                         & 53.4\%                                                          & 44.6\%                                                         & 31.6\%                                                             & 46.9\%                                                                & 37.3\%                                                                & 33.2\%        & 68.8\%        & 48.8\%         & 33.4\%                                                           & 40.2\%                                                             & 60.8\%                                                          & \cellcolor{gray!30}43.54\%                                                                   \\
$\tenx$@200TPP & 23.3\%                                                         & 53.4\%                                                          & 46.6\%                                                         & 31.2\%                                                             & 46.2\%                                                                & 38.8\%                                                                & 32.2\%        & 68.8\%        & 47.9\%         & 34.4\%                                                           & 38.4\%                                                             & 60.4\%                                                          & \cellcolor{gray!30}43.46\%                                                                   \\
$\dtoz$@200TPP & 24.7\%                                                         & 54.5\%                                                          & 48.2\%                                                         & 32.4\%                                                             & 50.0\%                                                                & 42.6\%                                                                & 32.9\%        & 70.1\%        & 50.4\%         & 32.6\%                                                           & 38.9\%                                                             & 62.5\%                                                          & \cellcolor{gray!30}45.00\%                                                                   \\ \bottomrule
\end{tabular}
\end{table}

\finding{Trends in validation loss also hold for \underline{downstream} evaluation.}

\cref{tab:downstream1} presents a variety of downstream
evaluations of the four models presented in
\cref{fig:train_loss}.
Differences between the models here are largely consistent with the
differences in training and validation loss, showing that $\dtoz$ is
meaningful not just for the autoregressive training objective, but for
real-world applications.

\subsection{Standard parameterization}\label{subsec:sp}

\begin{figure}
  \centering
  \scalebox{\onefigscale}{
  \includegraphics[width=\textwidth]{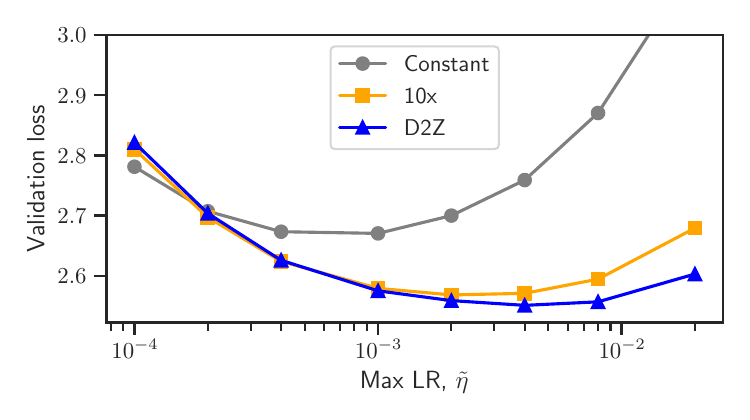}
  }
  \mbox{}
  \vspace{-4mm}
  \mbox{}
  \caption{\textbf{Standard parameterization results}:
    Validation loss for different LR and decay combinations, across
    peak LRs, for 610M-parameter models trained with the standard
    parameterization.\label{fig:maxlr_sp.curves}}
\end{figure}

\finding{Similar trends hold for the standard parameterization.}

\cref{fig:maxlr_sp.curves} presents results for a 610M-parameter
model trained with the standard parameterization.  Here $\hateta$ is
therefore not a $\mup$-corrected base LR, but rather a LR that we
swept directly for this model scale.
Results are obviously quite similar to results using $\mup$,
suggesting the benefits of $\dtoz$ are not $\mup$-specific.  Further
results using the standard parameterization, but for NanoGPT models,
are in \cref{subsec:nanogpt} below.

\subsection{Model sizes}\label{subsec:model_sizes}

\begin{figure}
  \centering
  \scalebox{0.95}{
  \includegraphics[width=\textwidth]{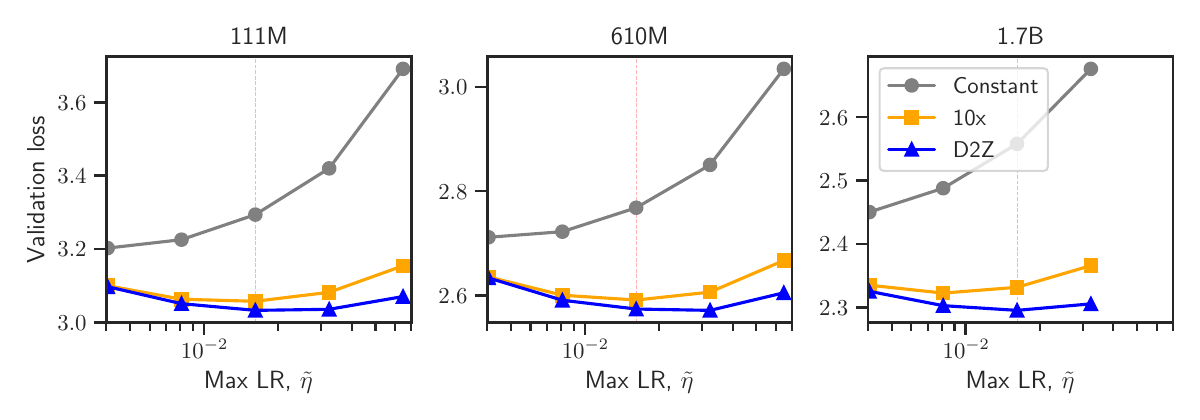}
  }
  \mbox{}
  \vspace{-4mm}
  \mbox{}
  \caption{\textbf{Model size comparison}: Validation losses at 20
    TPP\@.
    Across all model sizes, $\linear$-$\dtoz$ outperforms
    $\linear$-$\tenx$. Note: Missing high-LR values in all plots
    correspond to failed training runs due to \texttt{NaN}
    instabilities.\label{fig:maxlr_tpp:models}}
\end{figure}

\finding{Improvement of $\dtoz$ over $\tenx$ is greater at 1.7B
  scale.}

\cref{fig:maxlr_tpp:models} presents results across 111M, 610M,
and 1.7B model sizes, all trained to 20 TPP\@.  Note the absence of
results for the highest LR setting at the 1.7B-scale; at the very
highest LR, numerical instabilities led to failed training runs.
Otherwise, results are fairly similar across model sizes.  At the
proxy-tuned peak LR, the gap between $\dtoz$ and $\tenx$ is 0.81\%,
0.67\%, and 1.56\%, at the 111M, 610M, and 1.7B scales, respectively.
We further investigate the issue of whether the gap between $\dtoz$
and $\tenx$ varies with model size as part of our scaling law
experiments below (\cref{subsec:scaling}).

\subsection{TPP}\label{subsec:tpp}

\begin{figure}
  \centering
  \scalebox{0.95}{
  \includegraphics[width=\textwidth]{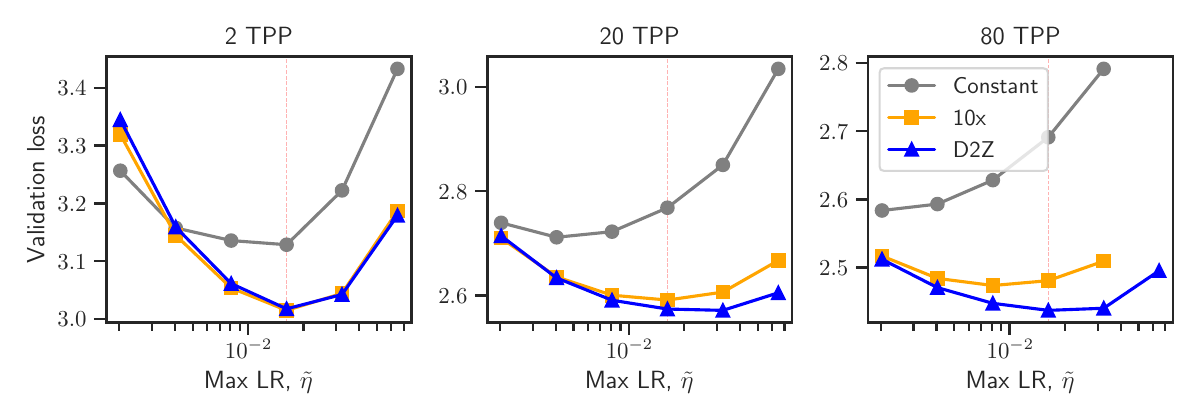}
  }
  \mbox{}
  \vspace{-1mm}
  \mbox{}
  \caption{\textbf{Comparing decay schedules across TPP} (610M
    scale):
    As TPP increases, $\linear$-$\dtoz$ begins to outperform
    $\linear$-$\tenx$, especially at the proxy-tuned peak LR (red
    lines).  The optimal LR also shifts significantly lower for
    $\constant$, somewhat lower for $\tenx$, and hardly at all for
    $\dtoz$.  Compare to \cref{fig:maxlr_tpp:111M} for 111M
    models.\label{fig:maxlr_tpp:610M}}
\end{figure}

\finding{Improvement of $\dtoz$ over $\tenx$ grows with TPP (610M scale).}

As we vary TPP, we consistently see increasing gains with $\dtoz$.
Here we plot the results for the 610M-scale models in
\cref{fig:maxlr_tpp:610M}, as a counterpart to main
\cref{fig:maxlr_tpp:111M}.

\subsection{Batch sizes}\label{subsec:batches}

\paragraph{Batch size setup}

Default batch sizes of 192 and 504 (\cref{tab:model_info}) were
selected for 111M and 610M scales.  These specific values were based
on an internal scaling law for optimal batch size as a function of
compute FLOPs (similar to those in~\citet{bi2024deepseek}
and~\citet{porian2024resolving}).  Batch size of 504 was then re-used
for 1.7B training (later testing confirmed this to be a good setting
at 1.7B for 20~TPP training).  In our experiments varying batch size,
we swept $B$ by factors of 2 around these initial settings.

\subsubsection{Iso-TPP batch size experiments}

\begin{figure}
\centering
\includegraphics[width=\textwidth]{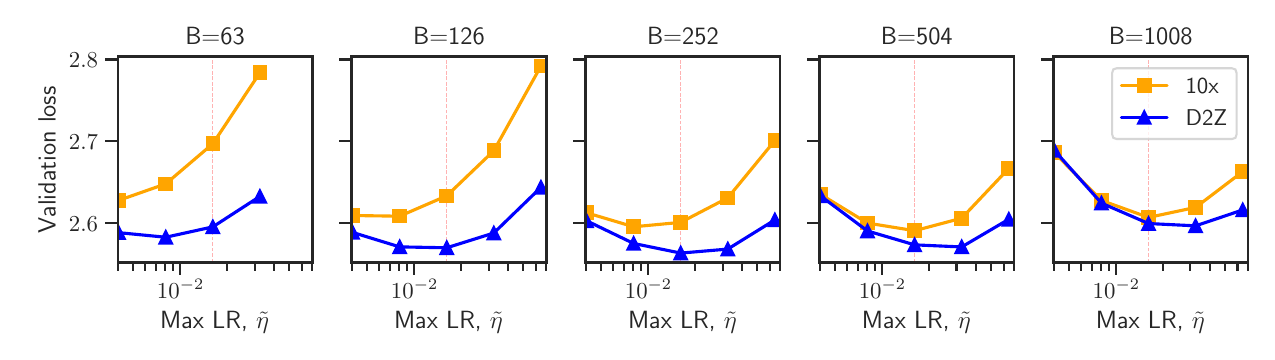}
\mbox{}
\vspace{-4mm}
\mbox{}
\captionof{figure}{\textbf{Comparison across batch sizes (iso-TPP)} (610M):
  Different \emph{batch sizes}, $B$, but \emph{all trained to 20~TPP}.
  As batch size decreases, the relative gain of $\dtoz$ over $\tenx$
  increases.  Same data as in \cref{fig:maxlr_isotppbatches}.
\label{fig:maxlr_isotppbatches_sep_batches}}
\end{figure}

\finding{Improvement of $\dtoz$ over $\tenx$ grows as batch size decreases (at constant TPP).}

In main paper \cref{fig:maxlr_isotppbatches}, we presented results
where all models train for the same number of total \emph{tokens}
(20~TPP), while only the batch size, $B$, varies.
\cref{fig:maxlr_isotppbatches_sep_batches} presents the same data, but
with each $B$ separated into a separate subplot; this lets us better
observe how the differences between $\dtoz$ and $\tenx$ evolve as $B$
changes.
We see clearly that for smaller $B$, as gradient noise increases, the
differences between $\dtoz$ and $\tenx$ also increase.  This again
demonstrates that LR decay is beneficial as a noise reduction
mechanism.

We now discuss the related observation that the optimal LR decreases
as $B$ decreases.  Note the EMA perspective of \citet{wang2024how}
would predict linear scaling of $\hateta$ with $B$.  This is because
the total number of iterations scales with $1/B$.  Therefore, in order
to keep the timescale constant (as a fraction of the total
iterations), we would need to scale $\hateta$ (or $\lambda$)
proportional to any change in $B$.
Although we do see roughly linear scaling in the case of $\tenx$
models, the optimal $\hateta$ is more stable with $\dtoz$.
In our conceptual model, the purpose of the timescale is to optimize
variance reduction.
But note that lowering $\hateta$ also has an effect on bias reduction.
For $\tenx$, as $B$ changes, it seems paying the cost of lower bias
reduction is worth the benefits in variance reduction.  Since using
$\dtoz$ already enables better variance reduction, it seems, for
$\dtoz$, lowering $\hateta$ has less value in the trade-off, so
optimal $\hateta$ lowers to a lesser extent.

\subsubsection{Iso-step batch size experiments} 

\begin{figure}
  \centering
  \includegraphics[width=\textwidth]{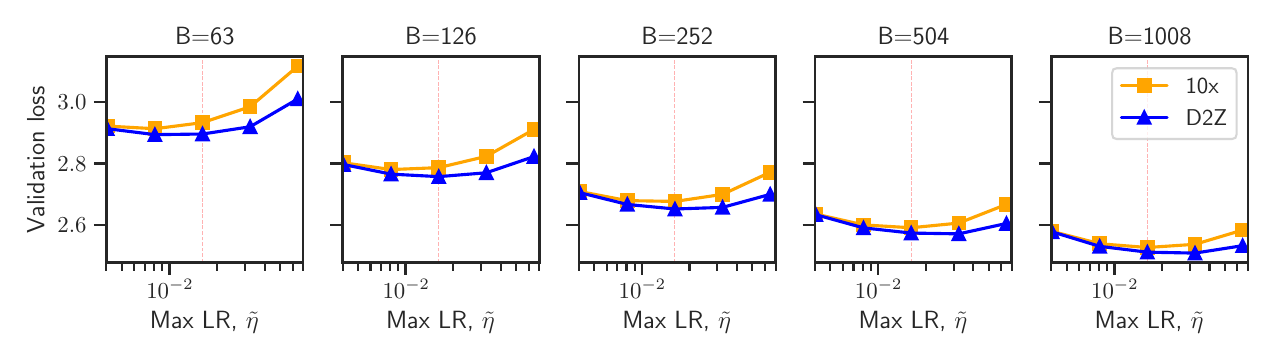}
  \mbox{}
  \vspace{-4mm}
  \mbox{}
  \caption{\textbf{Comparison across batch sizes (iso-step)} (610M):
    Different batch sizes, $B$, but all trained for \emph{11752 steps}
    (not iso-FLOP, smaller batches see fewer TPP).  In contrast,
    \cref{fig:maxlr_isotppbatches} and appendix
    \cref{fig:maxlr_isotppbatches_sep_batches} instead keep \emph{TPP}
    constant. $\dtoz$ remains superior in the iso-Step context.  More
    interestingly, here the optimal LR actually \emph{increases} as
    batches --- and, correspondingly, the \emph{datasets} --- increase
    in size.\label{fig:maxlr_isostepbatches}}
\end{figure}

\finding{Optimal LR \underline{increases} with batch size when
  training over the same number of batches/steps.}

In the above batch size tests, we kept the dataset size constant
(iso-TPP).  Now consider the following two mechanisms for increasing
the dataset size (TPP) of a training run:
\begin{enumerate}
\item Fix the batch size, but increase the number of steps.
\item Fix the number of steps, but increase the batch size.
\end{enumerate}
Approach (1) was taken in our experiments increasing TPP\@.  In that
case, we found optimal LR to decrease, as a mechanism for coping with
greater gradient noise, which grows with TPP\@.
We now discuss Approach (2): fixing the number of steps, but
increasing the batch size (iso-step).

In \cref{fig:maxlr_isostepbatches}, we keep the number of \emph{steps}
constant to 11752, so each model will see the same total number of
batches, and we increase $B$ by factors of two.\footnote{Note, for
purposes of scale, results at $B$=$504$ are the same in
\cref{fig:maxlr_isotppbatches_sep_batches} and
\cref{fig:maxlr_isostepbatches}; the latter just has a larger range on
the y-axis.}
Since number of steps does not change, \citet{wang2024how} do not
prescribe any adjustments to the timescale, and hence no adjustment is
needed to LR (or weight decay).
But we clearly see that optimal LR increases as $B$ increases
(like a rolling wheel, LR bowls rotate clockwise as we move to the
right).
This is evidently because increasing $B$ decreases gradient noise;
with less noise, we can afford a larger LR throughout training.
More precisely, increasing $\hateta$ decreases the number of AdamW
weight updates that are combined.  Therefore, an increase in $\hateta$
can be viewed as a way to reduce the effective number of batches that
are combined, basically compensating for the fact per-step $B$ is
larger.

As $B$ increases, gradient noise decreases, and we would expect
benefits of LR decay to also diminish; differences between $\tenx$ and
$\dtoz$ do seem to decrease.
This aligns with prior work, e.g., \citet{you2019how} compared SGD to
full-batch GD (with WideResNets for image classification), and showed
that when gradient noise is eliminated by using full GD, LR decay is
no longer beneficial.

%The deep connection optimal batch sizes of course also deeply depend on
%gradient noise~\citep{mccandlish2018empirical}.
%, it has a similar effect to decreasing the
%batch size

%Not to be confused with increasing the batch size and decreasing the
%number of steps, where \citet{wang2024how} would prescribe decreasing
%the timescale, so increasing the LR proportionally.  In this case, the
%number of iterations is constant, so \citeauthor{wang2024how} wouldn't
%advise doing anything specifically.
%
%Our results clarify that reducing $\alpha=\eta\lambda$ is primarily
%beneficial as a noise-reduction mechanism.
%
%Recall that in the view of \citet{wang2024how}, the benefit of
%$\alpha$ is to optimize the timescale of \emph{data}; if we double
%the amount of data, we should double the steps that we smooth over,
%e.g., \emph{decrease} $\eta$ or $\lambda$ by a factor of 2.
%%
%But consider \cref{fig:maxlr_isostepbatches}: as we double the amount
%of data (via doubling of the batch size and keeping the same number of
%steps), the optimal LR actually \emph{increases} rather than
%decreases.

\subsection{Weight Decay}\label{subsec:wd}

\subsubsection{Weight decay: results}
 
\finding{Benefits of weight decay are observed primarily when using LR
  $\dtoz$.}

\begin{figure}
  \centering
  \scalebox{\wdfigscales}{
    \includegraphics[width=\textwidth]{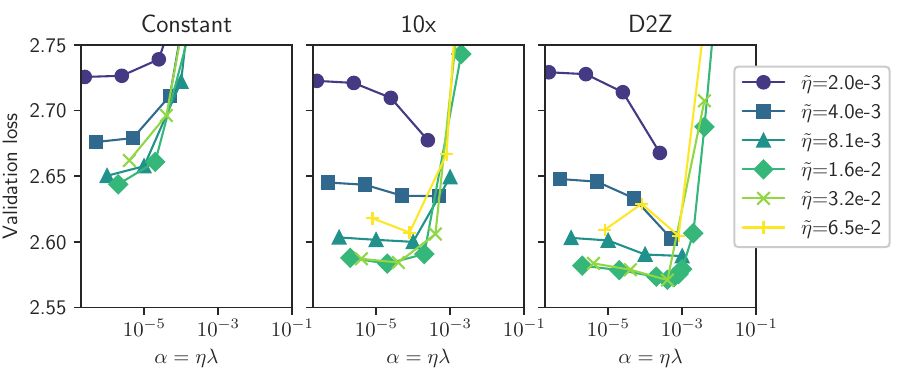}
  }
  \mbox{}
  \vspace{0mm}
  \mbox{}
  \caption{\textbf{Weight decay results, grouped by $\hateta$ as
      $\alpha$=$\eta\lambda$ varies} (610M, 20~TPP):
    Validation losses for different settings of decay schedule
    (separated by subplot), maximum learning rate ($\hateta$, marked
    by color), and weight decay ($\lambda$, corresponding to points on
    the LR curves).  Loss is plotted relative to
    $\alpha$=$\eta\lambda$.  The optimal loss corresponds to high (but
    not too high) $\hateta$.  As the smoothing $\alpha$ increases,
    $\constant$ models suffer, $\tenx$ models see marginal gains,
    while $\dtoz$ models benefit
    significantly.\label{fig:maxlr_wd:610M}}
\end{figure}

\cref{fig:maxlr_wd:610M} analyzes the interaction between \emph{weight
decay} (WD) and LR for 610M models.
Here we plot with the x-axis set to the (peak) smoothing parameter,
$\alpha$=$\eta\lambda$ (on a log scale), and y-axis set to validation
loss.  Note that for a given decay ratio and fixed $\alpha$, EMA
update coefficients (\cref{sec:emas}) are identical (i.e., specific
values of $\eta$ and $\lambda$ do not affect coefficients, only their
product).
Thus it supports the extended EMA perspective to note that regardless
of peak LR, $\hateta$, models tend to reach lowest loss at around the
same $\alpha$ (\cref{fig:maxlr_wd:610M}); that is, the same EMA
timescale, $\nicefrac{1}{\eta \lambda}$, is optimal across different
LRs.  However, to reach optimal loss, models require $\hateta$ to be
high enough, but not too high --- at $\hateta$=$6.5$e-$02$ and above,
we consistently encounter instabilities in training.

\begin{figure}
  \centering
  \scalebox{\wdfigscales}{
    \includegraphics[width=\textwidth]{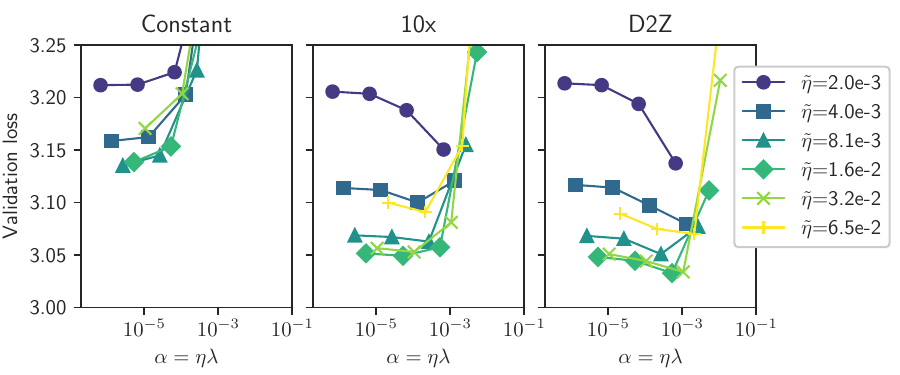}
  }
  \vspace{0mm}
  \caption{\textbf{Weight decay results, grouped by $\hateta$ as
      $\alpha$=$\eta\lambda$ varies} (111M, 20~TPP):
    Validation losses for different settings of decay schedule
    (separated by subplot), maximum learning rate ($\hateta$, marked
    by color), and weight decay ($\lambda$, corresponding to points on
    the LR curves).  Loss is plotted relative to
    $\alpha$=$\eta\lambda$.  The optimal loss is obtained when
    $\hateta$ is high, but not too high, typically around $\maxlr$ to
    $3.2$e-$02$.  As in \cref{fig:maxlr_wd:610M}, only $\dtoz$ models
    further improve as $\lambda$ increases.\label{fig:maxlr_wd:111M}}
\end{figure}

When increasing $\lambda$ with a given peak LR (moving left-to-right
along curves), note behavior differs depending on LR schedule:
$\constant$ models perform worse, $\linear$-$\tenx$ sees marginal
gains, and $\linear$-$\dtoz$ benefits significantly.  Results are
similar for 111M models (\cref{fig:maxlr_wd:111M}).

\begin{figure}
  \centering
  \scalebox{\wdfigscales}{
    \includegraphics[width=\textwidth]{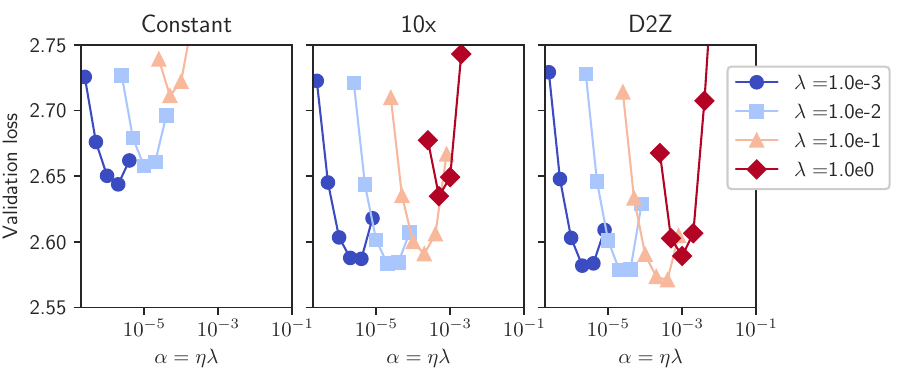}
    }
  \vspace{0mm}
  \caption{\textbf{Weight decay results, grouped by $\lambda$ as
      $\alpha$=$\eta\lambda$ varies} (610M, 20~TPP):
    Subset of the data in \cref{fig:maxlr_wd:610M}, but now curves
    trace points with same weight decay $\lambda$ (in color) and LR
    $\eta$ varies across each curve.  Only $\dtoz$ models
    significantly improve as we increase weight decay.  $\dtoz$ models
    are also less sensitive to choice of
    $\lambda$.\label{fig:maxlr_wdgroups:610M}}
\end{figure}

\begin{figure}
  \centering
  \scalebox{\wdfigscales}{
    \includegraphics[width=\textwidth]{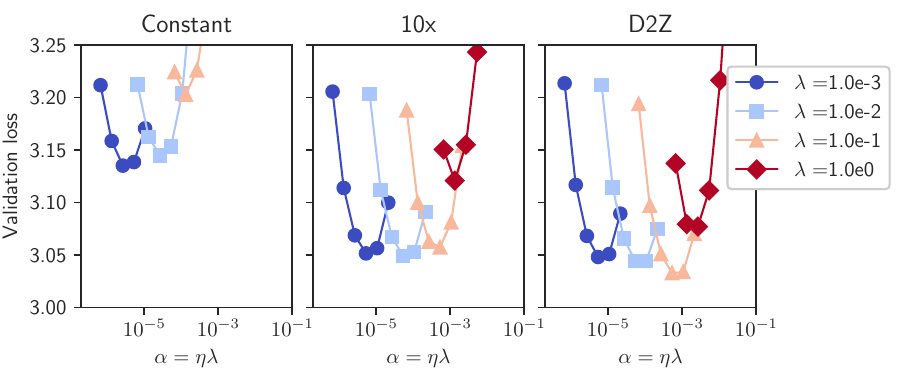}
    }
  \vspace{0mm}
  \caption{\textbf{Weight decay results, grouped by $\lambda$ as
      $\alpha$=$\eta\lambda$ varies} (111M, 20~TPP):
    Same data as in \cref{fig:maxlr_wd:111M}, but now curves trace
    points with same weight decay $\lambda$ (in color) and LR $\eta$
    varies across each curve.  Only $\dtoz$ models significantly
    improve as we increase weight decay.  $\dtoz$ models are also less
    sensitive to choice of $\lambda$.  See
    \cref{fig:maxlr_wdgroups:610M} for 610M-model
    results.\label{fig:maxlr_wdgroups:111M}}
\end{figure}

\cref{fig:maxlr_wdgroups:610M,fig:maxlr_wdgroups:111M} depict the same
data, but with points now grouped by $\lambda$.\footnote{Note
\cref{fig:maxlr_wdgroups:610M} excludes some points from
\cref{fig:maxlr_wd:610M}: \cref{fig:maxlr_wd:610M} includes some
additional $\lambda$ settings specifically for $\hateta$=$\maxlr$, and
we leave those off this figure in order to reduce clutter.} For higher
or lower $\lambda$, we cannot reach the optimal timescale without
taking $\hateta$ out of its comfort zone.  Therefore, with our 20 TPP
models, $\lambda$=0.1 emerges as good default setting because, in
synergy with a comfortable LR, $\lambda$=0.1 corresponds to an
appropriate EMA timescale.

Similar to the findings in \citet{andriushchenko2023why}, we also
confirm weight decay does not act like a traditional regularizer here:
beneficial settings of weight decay always improve both validation
\emph{and} training loss.

%TODO: note Loshchilov results here too.
%
% ``weight decay does not provide benefits when training with constant
% LRs as shown in Fig. 5 (right) emphasizing the importance of its
% interaction with LR decay'' -- I think we reach a similar
% explanation for this.

% although these works don't also optimize the peak LR.  Also,
% Andriushchenko believes Chinchilla compared WD to no WD, but I'm
% personally not so sure.  It just says they compare Adam to AdamW.
% But Adam *could* have a WD value.

%We also provide additional weight decay results.
%\cref{fig:maxlr_wdgroups:610M} is the same data as in
%\cref{fig:maxlr_wd:610M}, except we now group the points by weight
%decay $\lambda$ rather than peak LR $\hateta$ 

%\cref{fig:maxlr_wd:111M,fig:maxlr_wdgroups:111M}
%provide the 111M counterparts to the 610M-scale weight decay plots.
%Results are broadly similar.

\subsubsection{Weight decay: further discussion}

\paragraph{Learning rates reduce bias, weight decay best controls variance}

Recall that \cref{hyp:wdbias} made the observation that while LR
$\eta$ and weight decay $\lambda$ can both equally change the update
coefficients, only $\eta$ is effective in moving the model away from
initial conditions (due to the $\nicefrac{1}{\lambda}$ factor applied
to the updates).
This is confirmed by \cref{fig:maxlr_wd:610M,fig:maxlr_wd:111M}: only
when $\eta$ is sufficiently high do models achieve optimal loss.
Given such a high-enough $\eta$, the primary benefit of weight decay
is evidently to adjust $\alpha$=$\eta\lambda$ (and hence the timescale
$\nicefrac{1}{\alpha}$), to a setting that synergizes best with the
decay schedule (for 610M models, around $4$e-$04$).
Weight decay is specifically useful in this context because training
instabilities prevent reaching this optimal $\alpha$ purely through
increasing $\eta$.
% Just for your own sanity: you can't take a lower LR and increase WD,
% or you won't make initial progress.  You can't take a higher WD and
% decrease WD, as it would be unstable.
%
However, given $\eta$ and $\lambda$ both affect the timescale, $\alpha$=$\eta\lambda$
can be viewed as the \emph{effective} or
\emph{intrinsic} LR~\citep{li2020reconciling,wang2024how}, but only
later in training.

Since weight decay does not impact bias reduction, it follows that,
when the timescale needs to be adjusted at \emph{high TPP} due to the
greater gradient noise, it should be more effective to lower WD than
to lower LR, exactly as prescribed by \citet{wang2024how}.
In ongoing concurrent work, we do observe lower loss from tuning WD
compared to tuning LR\@.
This stands in contrast to other recent work that recommends purely
decreases in LR for high-TPP
training~\citep{shen2024power,bjorck2024scaling}.

\paragraph{Why weight decay is only beneficial when using LR decay}

Our observation that weight decay is only beneficial when using a LR
(decay) schedule aligns with both \citet{loshchilov2017decoupled} and
\citet{andriushchenko2023why} who also observed no benefit from WD
when using constant LRs.
In contrast, \citet{alephalpha2024introducing} recently reported
$\lambda$=$0.1$ to improve over $\lambda$=$0.0$ in LLM training;
notably, they train with $\dtoz$.
We believe the reason for these observations can also be explained by
the bias/variance trade-off, and by the fact, noted above, that weight
decay primarily plays a role later in training.
Since its LR is fixed, $\constant$ negotiates the bias/variance
trade-off by using a LR that is suboptimally \emph{low} early and
suboptimally \emph{high} later on (refer back to
\cref{fig:bias_var_all} in the main paper for a visualization).
Increasing $\lambda$ raises the already-too-high (late) $\alpha$ even
higher, further decreasing the timescale $\nicefrac{1}{\alpha}$ and
hurting the model.
In contrast, when using a LR schedule, weight decay can play its
beneficial variance reduction role, as discussed above.

It is also worth observing that if one is using $\dtoz$ with a LR that
is suboptimally low (e.g., if increasing it any further caused
numerical stability issues, as we observed with NanoGPT in
\cref{subsec:nanogpt}), then increasing weight decay can be very
beneficial.  For example, note the large improvements in loss as
weight decay is increased for the $\hateta$=$4.0$e-$3$ and
$\hateta$=$2.0$e-$3$ curves in \cref{fig:maxlr_wd:610M}.

\paragraph{Stability of $\dtoz$ loss with high $\lambda$}

%Why is LR more stable?
%
While $\constant$ and $\tenx$ are much worse at higher $\lambda$
values, $\dtoz$ performs reasonably well even at $\lambda$=$1.0$
(particularly at the 610M scale).  Since increasing $\lambda$
effectively decreases the timescale $\nicefrac{1}{\alpha}$ over which
weight updates are integrated, it increases the variance over weight
updates.  Therefore, setting $\lambda$ too high hurts all models, but
since $\dtoz$ has lower $\eta$ later in training, it somewhat
counterbalances the increase in $\lambda$ in the product
$\alpha$=$\eta\lambda$.  Another view of this is that as fewer updates
are combined, noise increases; $\dtoz$ is evidently best at mitigating
the negative effects of such noise through its maximal LR decay.

\subsection{Step decay}

\begin{figure}
  \centering
  \scalebox{1.0}{
  \includegraphics[width=\textwidth]{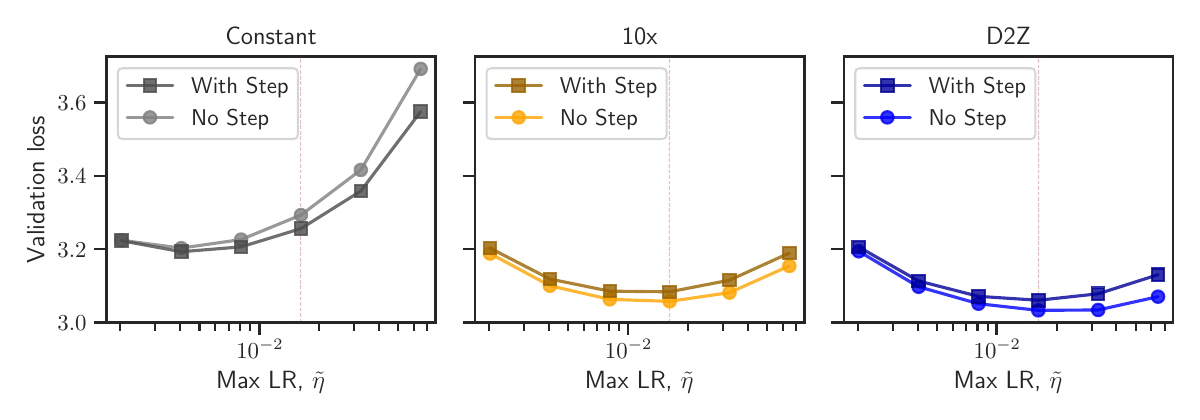}
  }
  \mbox{}
  \vspace{-2mm}
  \mbox{}
  \caption{\textbf{Impact of applying $\step$ decay} (111M, 20~TPP):
    Validation loss for different LR and decay combinations \emph{with
    and without $\step$ decay}.  $\step$ decay is applied after 90\%
    of training, stepping to a value equal to 0.1\% of the maximum LR\@.
    Dropping the LR in this manner helps $\constant$, although it is
    still below the level of $\tenx$ and $\dtoz$.  Adding stepping to
    a $\tenx$ or $\dtoz$ schedule is not
    beneficial.\label{fig:maxlr_step.curves}}
\end{figure}

\finding{Step decay helps $\constant$, but does not provide benefits
  if already decaying the LR.}

In \cref{fig:maxlr_step.curves}, we present results investigating the
impact of $\step$ decay on model training.  Here, for 111M-parameter
models, $\step$ decay can improve the loss versus keeping the LR
\underline{constant} (left panel), but the resulting losses are still
much worse than those obtained with $\dtoz$ or $\tenx$ decay.  We also
tried applying a $\step$ decay to a LR that had been following a
$\tenx$ (middle panel) or $\dtoz$ trajectory (right panel); this
approach always led to inferior results.

While it is likely possible to improve the quality of $\step$ decay by
tuning the positioning of the drop, we hypothesize that these efforts
will not surpass $\dtoz$, since dropping the LR will fundamentally
always result in an over-emphasis of earlier updates, as shown in
\cref{fig:background_ema} and \cref{fig:img_emas}.  Moreover,
introducing additional tunable hyperparameters (i.e., when and how
much to decay) is a further drawback of the $\step$ schedule.

%\paragraph{Inverse square root decay variant}

%Note that we always performed $\invsqr$ decay after performing our
%standard linear increase in the LR warmup.  This follows the approach
%in \citet{vaswani2017attention} and also
%\href{https://huggingface.co/docs/transformers/en/main\_classes/optimizer\_schedules\#transformers.get\_inverse\_sqrt\_schedule}{some
%  LLM libraries}).  Practitioners should be aware that warming up at a
%constant LR has also been explored~\citep{raffel2020exploring}, and is
%the approach used in
%\href{https://docs.cerebras.net/en/rel-2.3.0/wsc/api/cerebras\_pytorch/optim.html\#cerebras.pytorch.optim.lr\_scheduler.InverseSquareRootDecayLR}{other
%  libraries}).
%
%We also tried another kind of inverse square root decay, where the
%step is reset to zero for the decay phase, but this one drops too
%quickly.  This leads us to conclude that $\invsqrt$ is sensitive to
%the length of warmup (as we show in
%\cref{fig:more_ema_lrs}), which is not widely appreciated.

\subsection{Weight sparsity}\label{subsec:sparsity}

\finding{Relative improvement of $\dtoz$ over $\tenx$ \emph{increases} with weight sparsity.}

\paragraph{Weight sparsity setup}

In this section, we investigate the role of $\linear$-$\dtoz$ in the
context of models trained with \emph{unstructured weight sparsity}, a
promising direction for improving the efficiency of large neural
networks~\citep{hoefler2021sparsity}.
We parameterize with $\mup$'s sparse extension,
$\supar$~\citep{dey2024sparse}.  $\supar$ allows us to use the same
$\mup$ hyperparameters as with dense models, except we must now apply
corrections due to both model scaling (i.e., $\rho$,
\cref{subsec:mup}) and layer sparsity.
For these experiments, we sparsified all non-embedding layers of our
610M-parameter dense models by randomly fixing certain weights to zero
for the duration of training.
We trained all models for 11752 total steps using a batch size of 504,
i.e., the same amount of training data as we used for training the
corresponding 610M-parameter dense models to 20~TPP\@
(\cref{tab:model_info,tab:train_steps}).

%% Skip:
%%\input{fig_maxlr_s=0.9375.heatmap.tex}
\begin{figure}
  \centering
  \scalebox{\onefigscale}{
    \includegraphics[width=\textwidth]{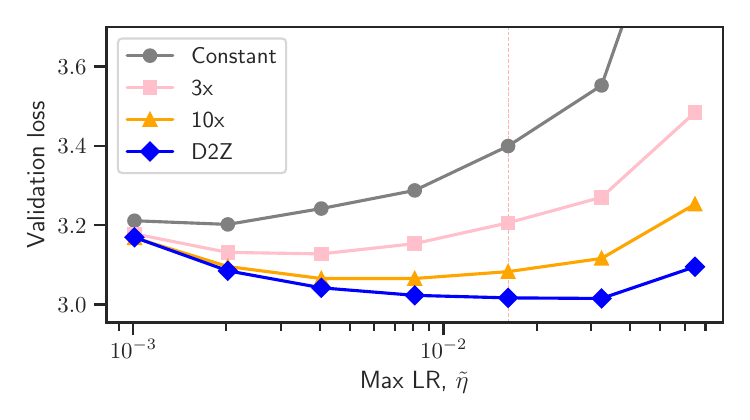}
  }
  \mbox{}
  \vspace{-4mm}
  \mbox{}
  \caption{\textbf{Comparison of decay schedules when using 93.75\%
      weight sparsity} (610M): Validation losses for different LR and
    decay combinations, for a model trained with $\supar$, with
    93.75\% unstructured weight sparsity.  As the decay rate
    increases, loss improves, while the optimal peak LR is much more
    stable when using $\dtoz$.\label{fig:maxlr_s=0.9375.curves}}
\end{figure}

At 93.75\% sparsity ($\nicefrac{1}{16}$ density), the optimal $\dtoz$
model improves by 1.64\% over the optimal $\tenx$ model, with a clear
trend of optimal LRs shifting lower and loss becoming worse as we go
from $\tenx$ to $3\times$ to $\constant$ decay
(\cref{fig:maxlr_s=0.9375.curves}).

\begin{figure}
\noindent % Ensures the minipages align with the left margin
\begin{minipage}[b]{0.49\textwidth}
  \centering
\input{fig_different_ratios.tex}
\end{minipage}%
\hfill % Use to fill space between the minipages
\begin{minipage}[b]{0.49\textwidth}
  \centering
  \input{fig_different_ratios_new.tex}
\end{minipage}
\end{figure}

At the proxy-tuned peak LR of $\maxlr$, $\dtoz$ is 2.15\% better than
$\tenx$.  We also trained 93.75\% sparse models with a variety of
other decay ratios between $\tenx$ and $\dtoz$, and present these
results in \cref{fig:different_ratios,fig:different_ratios_new}.
Here we see a largely linear decrease in loss with a linear increase
in the decay ratio (i.e., a linear decrease in the minimum LR).
These are encouraging findings in the sense that $\dtoz$ can seemingly
be used directly on a range of problems, without having to worry about
tuning a problem-specific LR decay ratio (e.g., 50$\times$ or
100$\times$).

\begin{figure}
  \centering
  \scalebox{\onefigscale}{
    \includegraphics[width=\textwidth]{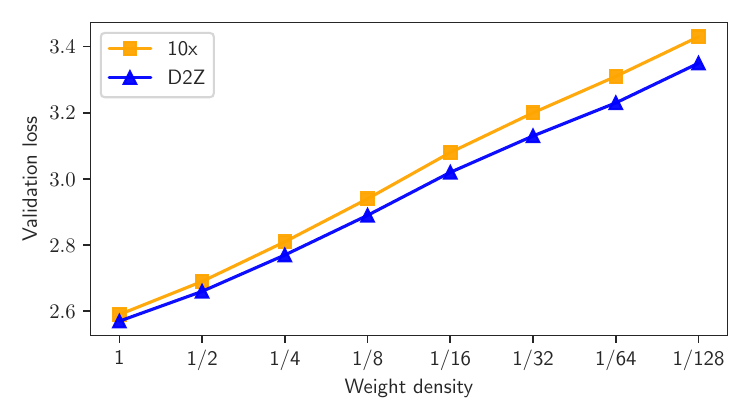}
  }
  \mbox{}
  \vspace{-1mm}
  \mbox{}
  \caption{\textbf{Comparison of decay schedules when using
      \emph{varying} weight sparsity} (610M): Validation losses for
    models trained with $\supar$, as unstructured weight sparsity
    increases (density decreases).  All models trained with the same
    number of training tokens (11752 steps, equivalent to 20~TPP for
    the fully-dense models).  As sparsity increases, the number of
    trainable parameters decreases, and thus
    tokens-per-(non-zero)-parameter (and the benefit of $\dtoz$)
    increases.\label{fig:cross_sparsity}}
\end{figure}

In \cref{fig:cross_sparsity}, we investigate the gap between
$\dtoz$ and $\tenx$ at the proxy-tuned peak LR across different sparsity
levels.  Note that increasing sparsity effectively leads to a
corresponding decrease in the number of trainable parameters.  Since
we use a fixed number of training tokens in each case, as the number
of parameters decreases, the number of tokens-per-parameter (TPP)
\emph{increases}.
In this sense, we note the relative differences between $\dtoz$ and
$\tenx$ are consistent with our results in \cref{fig:more_tpp}
--- as TPP (and gradient noise) increases, $\dtoz$ performs relatively
better.

\subsection{Error bars}

\finding{Run-to-run variation in results is low.}

\begin{figure}
  \centering
  \scalebox{\onefigscale}{
    \includegraphics[width=\textwidth]{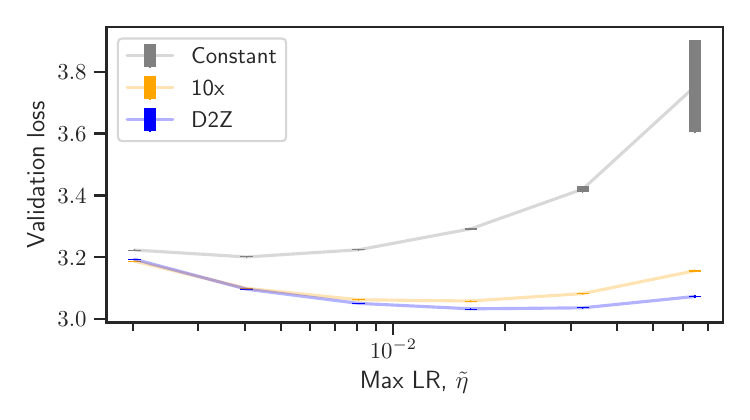}
  }
  \mbox{}
  \vspace{-1mm}
  \mbox{}
  \caption{\textbf{Stability across random seeds} (111M, 20~TPP):
    Validation loss variance for different maximum LRs at different
    decay ratios.  Each point corresponds to the mean validation loss
    over 5 separate training runs with different random seeds; the
    error bars give the standard deviation.  Beyond the instabilities
    at high learning rates, run-to-run variance is remarkably low at
    this scale.\label{fig:seeds}}
\end{figure}

Taken as a whole, our results are remarkably stable: empirical results
for different model scales, training durations, batch sizes, weight
decays, and weight sparsity settings largely behave as predicted by
theory.  Since all validation runs are performed on 1.1B tokens, any
significant run-to-run variance must arise during training.
To quantify this variance, we repeated 111M-model 20~TPP training four
additional times, resulting in 5 total validation loss results for
each original training run.  \cref{fig:seeds} confirms that
run-to-run variance is remarkably low.  The only significant variance
arises in $\constant$ results at the highest learning rate.  Here, the
training loss sometimes spikes at various points during training, and
the final validation loss can be significantly higher for models that
cannot recover sufficiently following the spike.

\subsection{NanoGPT experiments}\label{subsec:nanogpt}

\finding{Improvement of $\dtoz$ over $\tenx$ persists in NanoGPT
  models --- when numerical issues are solved.}

We also compared $\tenx$ versus $\dtoz$ by training NanoGPT models
using the NanoGPT codebase \citep{karpathy2024nanogpt}.
NanoGPT uses the standard parameterization.  We configured these
models to be largely similar to our 111M-parameter models, also using
a weight decay of 0.1, a context length of 2048, and the GPT-2 vocab
size of 50257.  Key differences here are that we do not include bias
weights, and we trained on the OpenWebText dataset
\citep{gokaslan2019owt}.
Experiments are run on Nvidia A10 GPUs.

As mentioned in \cref{subsec:confounding}, we initially tested
NanoGPT in \verb|bfloat16| precision.  Here, at the default NanoGPT
learning rate of 6e-4, $\tenx$ performed slightly better than $\dtoz$.
As we pushed the LR 50\% higher (to 9e-4), both $\tenx$ and $\dtoz$
had higher loss, and by 1.2e-3, the loss from $\tenx$ doubled.

\begin{figure}
  \centering
  \scalebox{\onefigscale}{
  \includegraphics[width=\textwidth]{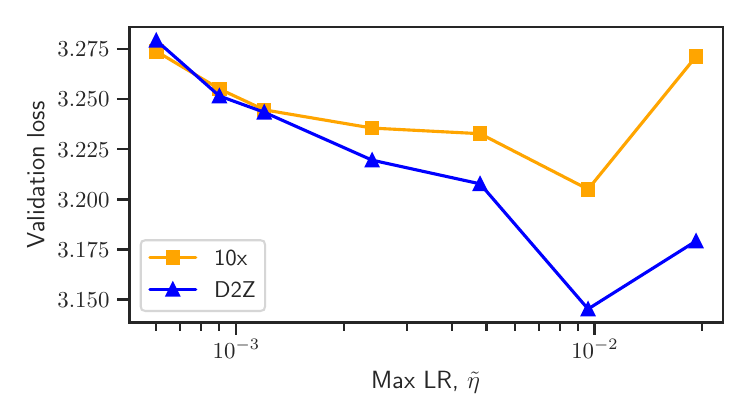}
  }
  \mbox{}
  \vspace{-1mm}
  \mbox{}
  \caption{\textbf{NanoGPT results} (111M, 20~TPP): Validation loss
    for different LR and decay combinations, for a \emph{NanoGPT
    model}.  With \texttt{float16} precision, we were not able to
    train above $6$e-$04$ without instabilities (first point in
    curves).  Moreover, at $\hateta = 6$e-$04$, $\tenx$ performed
    better than $\dtoz$.  After switching to \texttt{float32}, we were
    able to train at higher $\hateta$ values, where $\dtoz$
    demonstrates its familiar superiority over
    $\tenx$.\label{fig:maxlr_nanogpt_curves}}
\end{figure}

We suspected that numerical issues may be causing the instabilities,
and repeated our experiments in \verb|float32|.  At this precision, we
were able to successfully increase the LR, by factors of two, up to 32
times the default.
At these levels, we do see the familiar gains of $\dtoz$ over $\tenx$
(\cref{fig:maxlr_nanogpt_curves}).
We note there is nothing fundamentally limiting about \verb|float16|
precision itself - indeed all our main experiments were done using
this precision.  Rather, \verb|float16| is simply problematic at a
high learning rate in the NanoGPT codebase (see \citet[Appendix
  C.2]{gray2024normalization} for a potential root cause of this
instability).

These experiments demonstrate that a comparison between $\dtoz$ and
$\tenx$ may serve as a kind of \emph{diagnostic} of whether a model is
being trained at optimal peak LRs: if a 100M+ model trained to 20~TPP
does not see roughly 1\% gains from using $\dtoz$, it is likely the LR
is not high enough.  In order to raise the LR further, efforts to
stabilize the model, perhaps including $\mup$ or other
techniques~\citep{wortsman2023small}, may be warranted.

\subsection{Scaling law experiments}\label{subsec:scaling}

\finding{Improvement of $\dtoz$ over $\tenx$ grows across model scales, for a different model and training setup.}

Encouraged by the results of $\dtoz$ at smaller scales, we began
testing $\dtoz$ in some of our frontier model efforts.  Frontier
models do not provide scope for hyperparameter tuning at scale.  Thus
it becomes important to derive scaling laws to forecast loss at larger
scales, based on the loss with a sequence of smaller models.

For this set of experiments, we tested
Llama-style~\citep{touvron2023llama} architectures, except using
LayerNorm instead of RMSNorm, and multi-head attention instead of
group-query attention.  We use $\mup$ here as well, and a batch size
scaling law to determine an optimal batch size for each model scale.
These models also use ALiBi embeddings~\citep{press2022alibi} and
SwiGLU~\citep{shazeer2020glu}.  Here, the context length is 8192
tokens, and we use tied embeddings.
We also re-tuned our $\mup$ proxy-model hyperparameters.

We used a bilingual data mix of English, Arabic and source code
samples mixed in a 2:1:0.4 mix ratio. English data is from
the~Pile~\citep{gao2020pile}, Arabic uses a proprietary
dataset~\citep{sengupta2023jais}, and the source code comes from the
GitHub portion of the Pile.

\begin{table}
  \centering
  \caption{\textbf{Model architecture and batch sizes for scaling law experiments}.\label{tab:model_info_scaling}}
\begin{tabular}{@{}ccccccc@{}}
\toprule
Model & vocab. size & $\dmodel$ & $\nlayers$ & $\dhead$ & $\dffn$ & batch size \\ \midrule
272M  & 84992       & 1024      & 14         & 64       &  2912   & 128  \\
653M  & 84992       & 1536      & 18         & 128      &  4096   & 160  \\
1.39B & 84992       & 2048      & 24         & 128      &  5472   & 208  \\
2.75B & 84992       & 2560      & 32         & 128      &  6832   & 256  \\ \bottomrule
\end{tabular}
\end{table}

To derive scaling laws to compare $\dtoz$ and $\tenx$ models, we used
the power law functional form $y = c x^m$, where $x$ is the
pre-training FLOPs, $y$ is the loss on the Pile validation set, and
$c$ and $m$ are parameters to be fit.
To fit these parameters, we transformed our Loss-to-FLOPs equation, $y
= c x^m$, to a logarithmic form, $log(y)=log(c)+m log(x)$, and fit the
slope and intercept parameters of this line using a least-squares
linear regression.
We trained models at four sizes to compute-optimal 20~TPP
(\cref{tab:model_info_scaling}), and computed total FLOPs spent
as well as validation loss on the Pile.
We then fit the power law free parameters to obtain our scaling laws.

\begin{figure}
  \centering
  \scalebox{1.0}{
    \includegraphics[width=\textwidth]{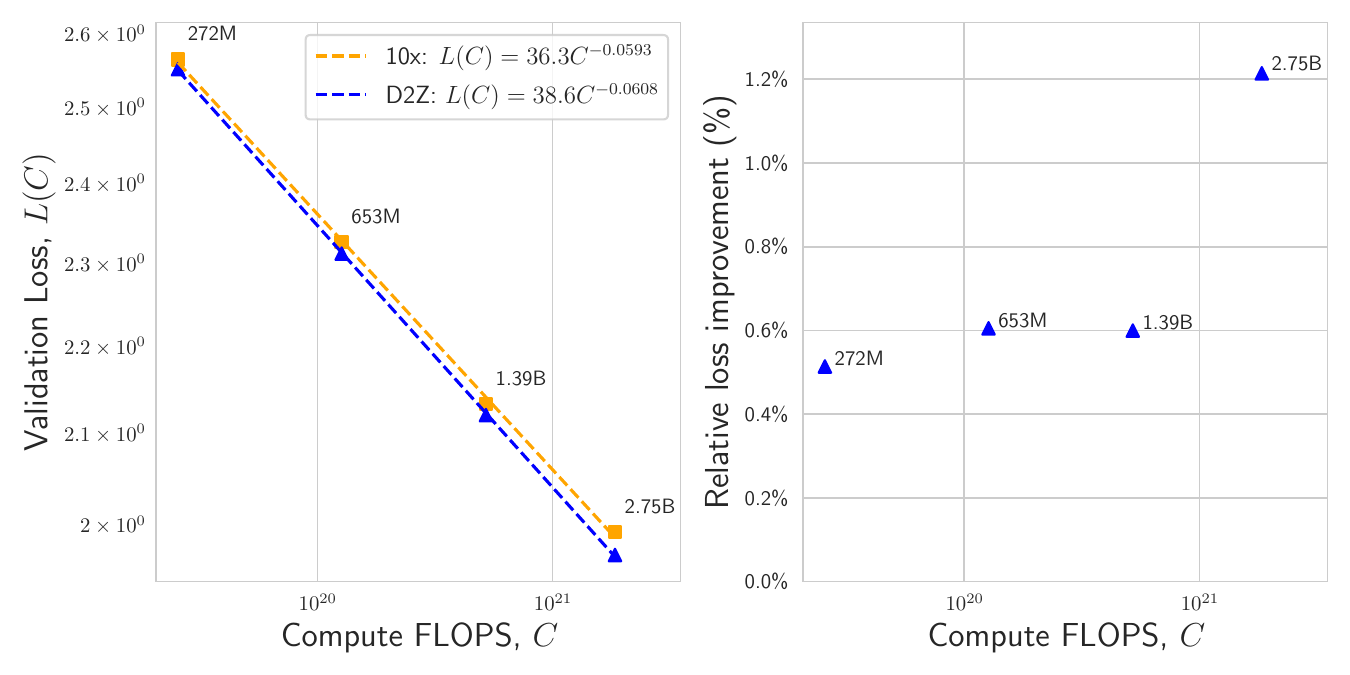}
  }
  \mbox{}
  \vspace{-2mm}
  \mbox{}
  \caption{\textbf{Scaling law comparison of decay schedules}:
    Loss-to-FLOPS scaling law fits for models trained to 20~TPP (8192
    context lengths, Llama architecture, Pile validation data).  The
    power law fit for $\dtoz$ models has a steeper slope than the
    scaling law for $\tenx$ models, implying a growing performance gap
    for larger model sizes.\label{fig:pile_scaling}}
\end{figure}

Encouragingly, here we find the scaling law slope of $\dtoz$ is
roughly 2.5\% better than $\tenx$ decay
(\cref{fig:pile_scaling}).
This translates to an improvement of roughly 1\% at 1.3B and 2.7B
scales, broadly similar to our earlier results at 1.7B scale.
Projecting our scaling law to a 70B model trained to a compute optimal
20~TPP, $\dtoz$ would achieve a roughly 2\% loss improvement over
$\tenx$ decay.
% If we trained such a model to higher TPP in order to enable
% efficient inference, we suspect gains would multiply

\section{AdamW as convex combination of weight updates, driven by LR schedule: Further details}

We have begun using plots of the weight update coefficients internally
to proactively analyze LR schedules.  These plots give us insight into
the timescale over which training data is integrated into the model
parameters, and can be obtained without actually performing any
training.
Moreover, we can also use the extended EMA perspective in order to
design novel LR schedules that have desirable blends of weight
updates.  Optimizing these coefficients thus provides a dual to
optimizing the LR decay function, which is why we refer to these
coefficients as the \emph{dual coefficients} of the LR schedule.
In this section, we first describe the design of one such LR schedule,
and then we follow with example plots for this and other schedules.
%
%This formulation enables proactive analysis of existing LR schedules
%--- by calculating coefficients and judging how updates are integrated
%over training --- without needing to actually train.  For example,
%Figure~\ref{fig:background_ema} (Section~\ref{sec:intro}) compares the
%effect of $\linear$, $\cosine$, and $\step$ decay on the output at
%step $t$ of a 111M-parameter model trained for 200 TPP (2D plots of
%$c_{t,i}$ over all $t$ and $i$ are in appendix
%Figure~\ref{fig:img_emas}; see also Figures~\ref{fig:more_ema_lrs}
%and~\ref{fig:inf_ema_lrs}).

%To what extent do the bias and variance terms play a role in modern
%LLM training?  We hypothesize that the answer to this question is
%\emph{scale-invariant} given a fixed training tokens-per-parameter
%(TPP).  We know that, regardless of scale, models tend to reach the
%compute-efficient-training frontier at around
%20~TPP~\citep{hoffmann2022empirical}.  It seems unlikely that at some
%scales, models train efficiently to 20~TPP by mostly minimizing bias,
%while at others they train efficiently to 20~TPP by mostly grappling
%with variance.  It is more likely that 20~TPP is consistently
%efficient across scales precisely \emph{because} it corresponds to a
%balanced combination of bias and variance.
% models are trained in a compute-efficient manner.  That is, provided
%we train models so that either training FLOPs or inference FLOPs are
%minimized (corrsponding to 20~TPP following Chinchilla scaling
%laws~\citep{hoffmann2022empirical}),

\subsection{Truly schedule-free LR schedules}\label{subsec:rational}

Recall that, for a moving average with time-varying smoothing,
$y_t = (1 - \alpha_t)y_{t-1} + \alpha_t x_t$,
we use
$c_{t,i} = \left( \prod_{j=i+1}^{t} (1 - \alpha_j) \right) \alpha_i$
to denote the dual coefficients,
i.e., the contribution of input $x_i$ to output $y_t$ at time $t$.
$\constant$ schedules, or schedules with a long constant phase such as
$\wsd$, are not truly ``schedule-free'' because their peak LR setting
affects the $(1 - \alpha_j)$ terms in the dual.  Different LRs will
correspond to different effective timescales over updates, and thus
different LRs will be optimal for different training
durations.\footnote{The different shapes of the coefficients for
different LRs can be seen in \cref{fig:more_ema_lrs:constant} for
$\constant$ and \cref{fig:inf_ema_lrs:wsd} for $\wsd$.}

In contrast, we now derive a schedule such that coefficients are
always weighted equally, at every training step.  First, it can be
shown that:
\begin{equation}
    \frac{c_{t,i+1}}{c_{t,i}} = \frac{\alpha_{i+1}}{(1-\alpha_{i+1})\alpha_i}.
\end{equation}
For the coefficients to be uniform at any step $t$, we require this
ratio to be 1.  Unity of this ratio implies that smoothing evolves
$\alpha_{i+1}$ = $\frac{\alpha_i}{(1+\alpha_i)}$.  Assuming a fixed
weight decay (so $\alpha_i = \eta_i\lambda$), coefficients will be
equal if the LR evolves:
\begin{equation}\label{eqn:rational}
  \eta_{i+1} = \frac{\eta_i}{(1+\eta_i\lambda)}
\end{equation}
We can initialize $\eta_0$ to some value and iterate
\cref{eqn:rational} to generate the full LR schedule.  We call this
the $\rational$ schedule since it is both a rational expression of
$\eta_i$ and a very reasonable approach: regardless of how long we
train, all weight updates contribute equally.  At each step we
effectively decrease all prior coefficients such that they now equal
the coefficient of the current update, $\alpha_t$.
%
% This may minimize catastrophic forgeting.

With $\lambda$=$1$ and no warmup, this schedule evolves like
$\nicefrac{1}{n}$.  However, in order to distance ourselves from the
initial conditions, we can warmup the LR in the usual manner to the
desired peak LR, then switch on rational decay, ensuring equal
coefficients going forward.  The full schedule is depicted in
\cref{fig:more_ema_lrs:rational}.
However, such a schedule will almost surely not work as effectively as
$\linear$ $\dtoz$ for fixed-duration training, since it lacks the
cooldown phase where gradient noise is minimized.  Indeed,
$\nicefrac{1}{n}$ has performed relatively poorly in prior
studies~\citep{ge2019step,defazio2023when}.
However, we introduce it here as a promising schedule for
\emph{continuous} pre-training.\footnote{In fact, \nicefrac{1}{n}
decay has long been regarded as optimal for strongly-convex
loss~\citep{robbins1951stochastic}, and optimal in other contexts when
combined with \emph{averaging}~\citep{defazio2023when}.  Since
averaging is an alternative to cooldown~\citep{hagele2024scaling},
$\rational$ plus averaging is a compelling schedule-free direction.}
%
%The implication is: no matter how long we train for, we will always
%incorporate all the knowledge gained previously.  No catastrophic
%forgetting.

%Fascinating: ``The Robbins-Monro conditions are the foundation of
%early learning rate theory (Robbins and Monro, 1951).  They advocated
%for step size sequences where ... of the schedules satisfying these
%conditions, they advocated for schedules with 1/t decay as they are
%asymptotically optimal for twice continuously differentiable and
%strongly convex functions. This schedule was later shown to be optimal
%(even non-asymptotically) for strongly convex G-Lipschitz stochastic
%optimization when appropriate averaging is used (Shamir and Zhang,
%2013; Lacoste-Julien et al., 2012; Rakhlin et al., 2012).''  So
%averaging takes the place of cooldown, and this is apparently optimal
%-- this is exactly what you might have thought!

\subsection{LR curves and dual coefficients}

\begin{figure}
  \centering
  \makebox[\textwidth][c]{
      \begin{subfigure}{\shrinkfigimgs\textwidth}
        \includegraphics[trim={0.28cm 0cm 0.3cm 0.35cm}, clip, width=\textwidth]{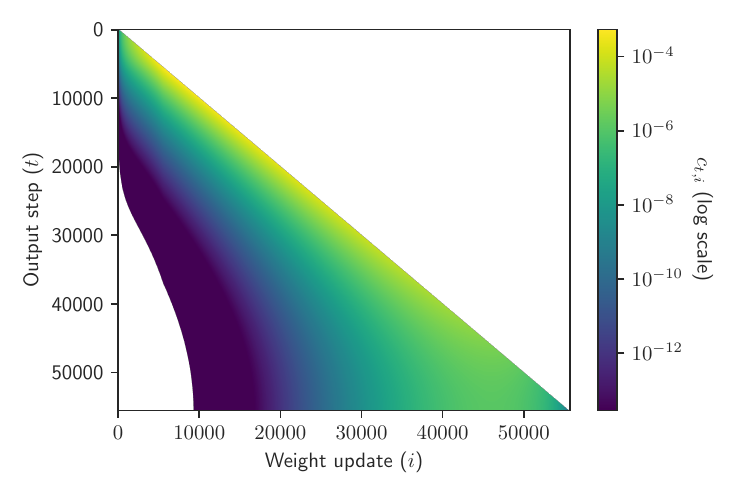}
        \subcaption{$\linear$}
        \label{fig:img_emas:d2z}
      \end{subfigure}
      \hspace{-1mm}
      \begin{subfigure}{\shrinkfigimgs\textwidth}
        \includegraphics[trim={0.28cm 0cm 0.3cm 0.35cm}, clip, width=\textwidth]{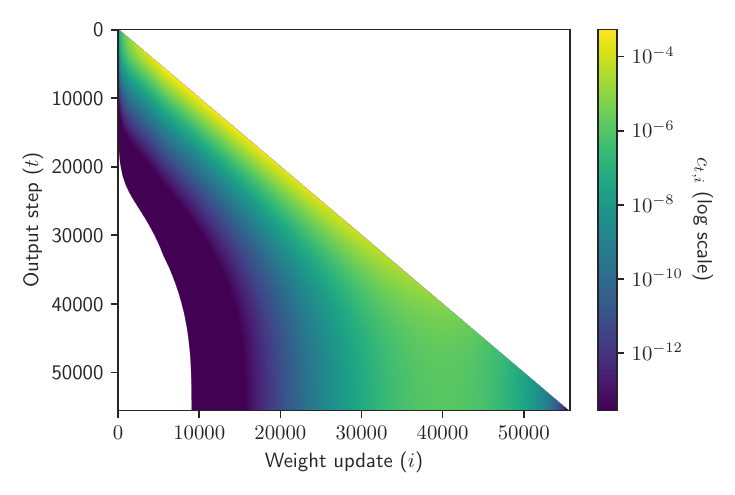}
        \subcaption{$\cosine$}
        \label{fig:img_emas:cosine}
      \end{subfigure}
  }
  \mbox{}
  \vspace{-2mm}
  \mbox{}
  \makebox[\textwidth][c]{
      \begin{subfigure}{\shrinkfigimgs\textwidth}
        \includegraphics[trim={0.28cm 0cm 0.3cm 0.35cm}, clip, width=\textwidth]{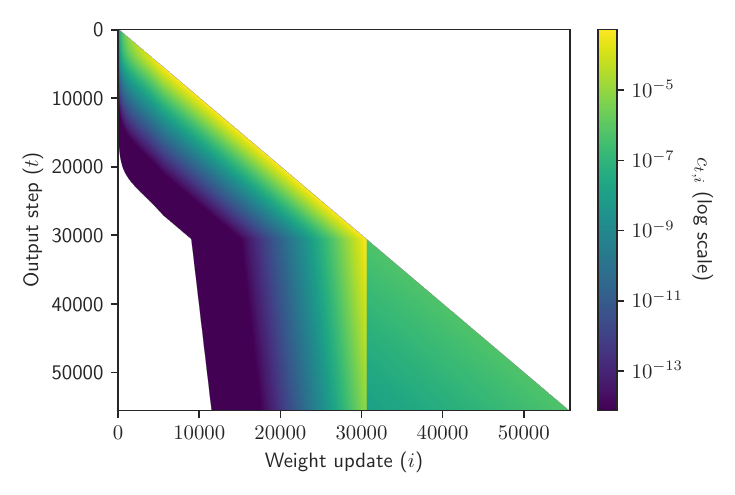}
        \subcaption{$\step$}
        \label{fig:img_emas:step}
      \end{subfigure}
      \hspace{-1mm}
      \begin{subfigure}{\shrinkfigimgs\textwidth}
        \includegraphics[trim={0.28cm 0cm 0.3cm 0.35cm}, clip, width=\textwidth]{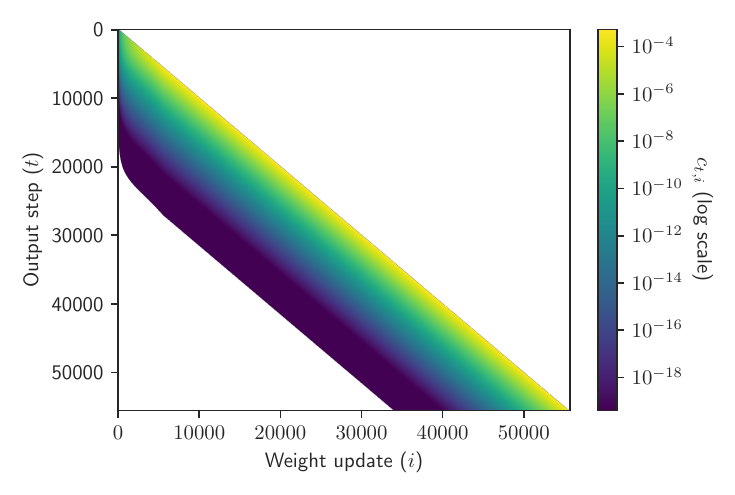}
        \subcaption{$\constant$}
        \label{fig:img_emas:constant}
      \end{subfigure}
  }
  \mbox{}
  \vspace{-4mm}
  \mbox{}
  \caption{\textbf{EMA perspective: weight update contribution across
      steps}: Convex combination of weight updates, with color
    indicating value of combination coefficients $c_{t,i}$: each
    $c_{t,i}$ gives the contribution of the $i$th update (across
    x-axis) to model weights $\theta_t$ across steps $t$ (y-axis).
    Note that LR schedules and coefficients corresponding to the final
    step only were presented earlier in \cref{fig:background_ema}
    (except for $\constant$).
    Coefficients correspond to settings for 111M-parameter models trained
    to 200~TPP: $t$=$55680$, $\hateta$=$\maxlr$,
    $\rho$=$\nicefrac{1}{3}$, $\lambda$=$0.1$.\label{fig:img_emas}}
\end{figure}

\begin{figure}
  \centering
  \makebox[\textwidth][c]{
    \begin{minipage}{\textwidth}
      \begin{subfigure}{\shrinkfigtwo\textwidth}
        \includegraphics[width=\textwidth]{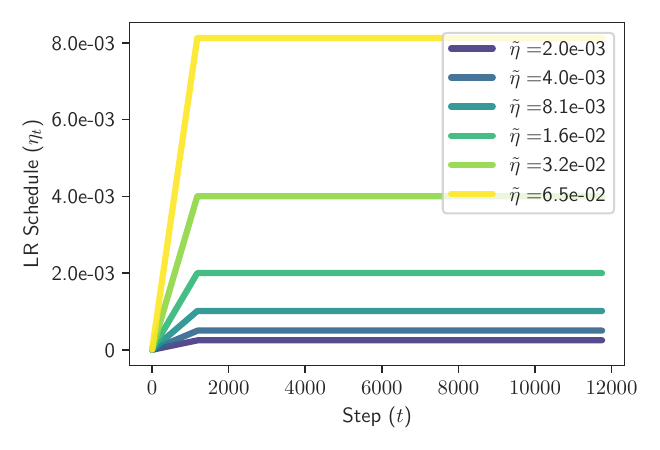}
      \end{subfigure}
      \hspace{-1mm}
      \begin{subfigure}{\shrinkfigtwo\textwidth}
        \includegraphics[width=\textwidth]{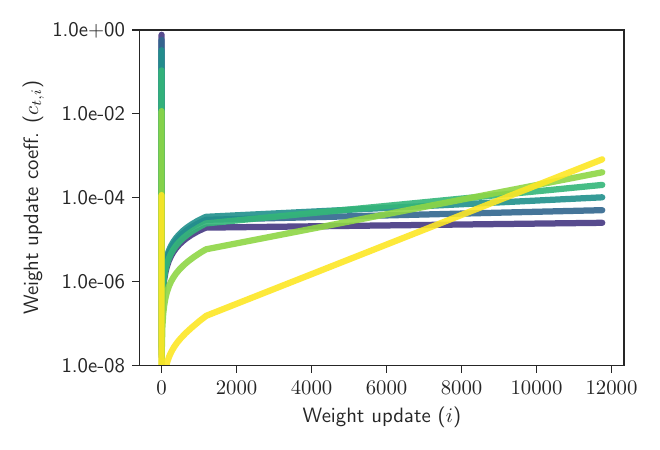}
      \end{subfigure}
      \vspace{-6.3mm}
      \subcaption{$\constant$}
      \label{fig:more_ema_lrs:constant}
    \end{minipage}
  }
  \mbox{}
  \vspace{0.5mm}
  \mbox{}
  \makebox[\textwidth][c]{
    \begin{minipage}{\textwidth}
      \begin{subfigure}{\shrinkfigtwo\textwidth}
        \includegraphics[width=\textwidth]{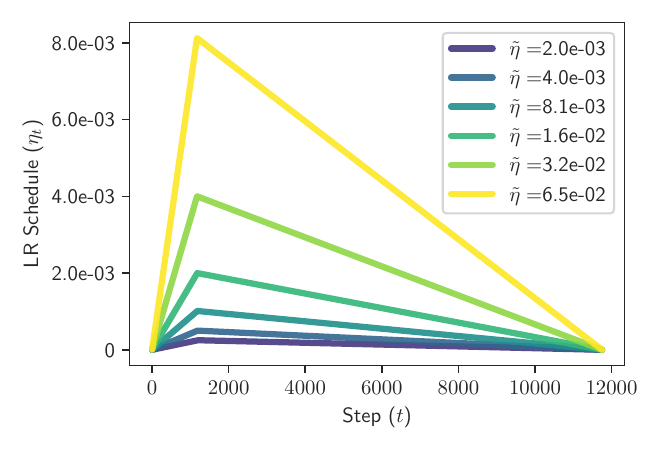}
      \end{subfigure}
      \hspace{-1mm}
      \begin{subfigure}{\shrinkfigtwo\textwidth}
        \includegraphics[width=\textwidth]{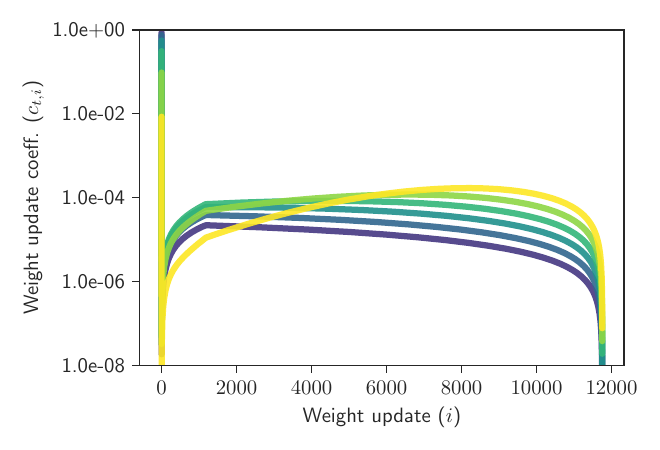}
      \end{subfigure}
      \vspace{-6.3mm}
      \subcaption{$\linear$ $\dtoz$}
      \label{fig:more_ema_lrs:d2zs}
    \end{minipage}
  }
  \mbox{}
  \vspace{0.5mm}
  \mbox{}
  \makebox[\textwidth][c]{
    \begin{minipage}{\textwidth}
      \begin{subfigure}{\shrinkfigtwo\textwidth}
        \includegraphics[width=\textwidth]{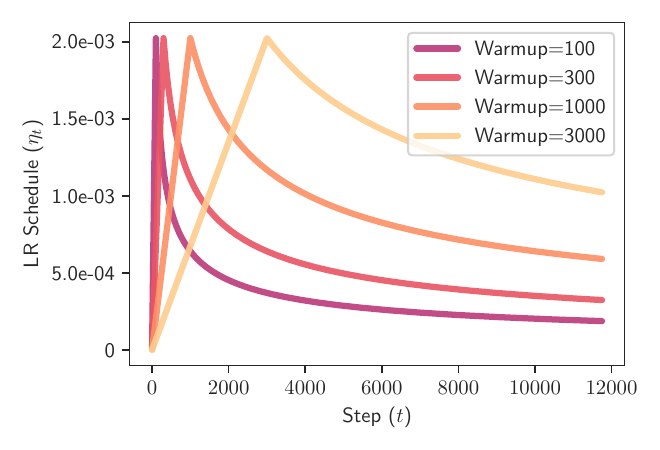}
      \end{subfigure}
      \hspace{-1mm}
      \begin{subfigure}{\shrinkfigtwo\textwidth}
        \includegraphics[width=\textwidth]{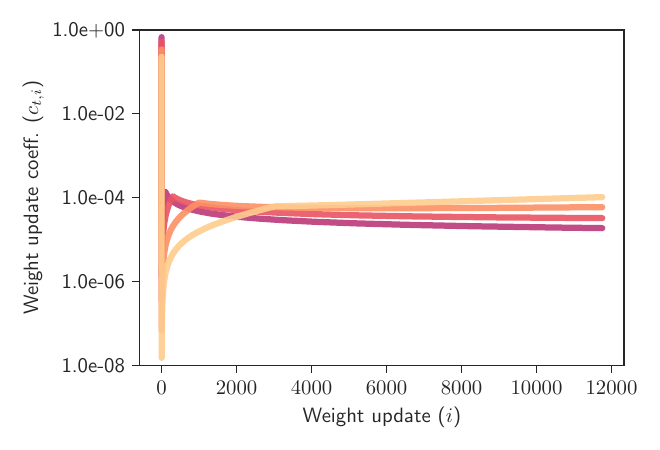}
      \end{subfigure}
      \vspace{-6.3mm}
      \subcaption{$\invsqrt$}
      \label{fig:more_ema_lrs:invsqrt}
    \end{minipage}
  }
  \mbox{}
  \vspace{0.5mm}
  \mbox{}
  \makebox[\textwidth][c]{
    \begin{minipage}{\textwidth}
      \begin{subfigure}{\shrinkfigtwo\textwidth}
        \includegraphics[width=\textwidth]{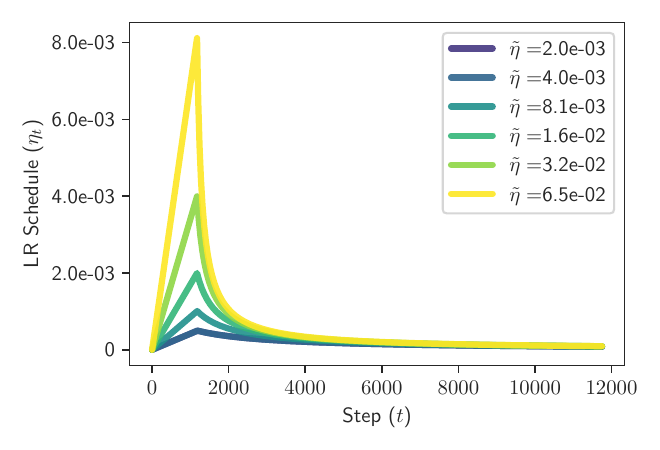}
      \end{subfigure}
      \hspace{-1mm}
      \begin{subfigure}{\shrinkfigtwo\textwidth}
        \includegraphics[width=\textwidth]{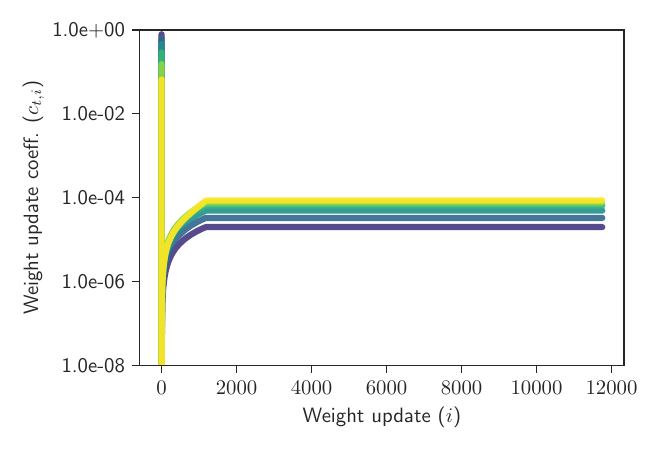}
      \end{subfigure}
      \vspace{-6.3mm}
      \subcaption{$\rational$}
      \label{fig:more_ema_lrs:rational}
    \end{minipage}
  } \mbox{}
  \vspace{0.5mm}
  \mbox{}
  \caption{\textbf{LR curves and dual coefficients for various common
      LR schedules} as well as the proposed $\rational$ approach
    (\cref{subsec:rational}): Dual coefficients shown at final
    training step (right side) for 610M-parameter, 20~TPP training
    ($t$=$11752$, $\rho$=$\nicefrac{1}{8}$, $\lambda$=$0.1$).  For
    $\invsqrt$, we vary the warmup and fix $\hateta$=$\maxlr$ for all
    curves.  Note coefficient values are plotted on
    log-scale\label{fig:more_ema_lrs}}
\end{figure}

\begin{figure}
  \centering
  \makebox[\textwidth][c]{
    \begin{minipage}{\textwidth}
      \begin{subfigure}{\shrinkfigtwo\textwidth}
        \includegraphics[width=\textwidth]{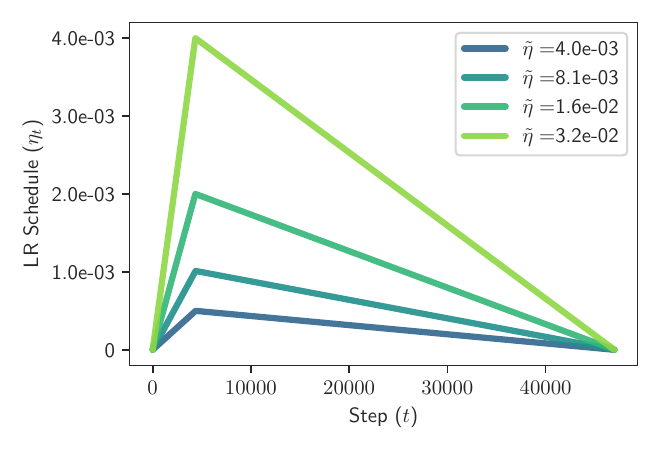}
      \end{subfigure}
      \hspace{-1mm}
      \begin{subfigure}{\shrinkfigtwo\textwidth}
        \includegraphics[width=\textwidth]{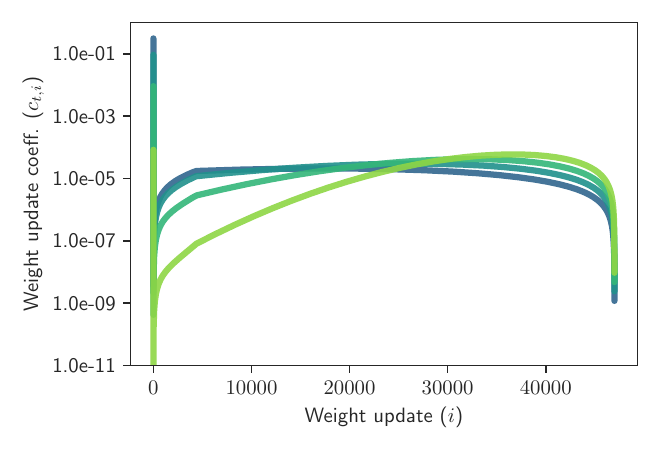}
      \end{subfigure}
      \vspace{-6.3mm}
      \subcaption{$\linear$ $\dtoz$}
      \label{fig:inf_ema_lrs:linear}
    \end{minipage}
  }
  \mbox{}
  \vspace{0.5mm}
  \mbox{}
  \makebox[\textwidth][c]{
    \begin{minipage}{\textwidth}
      \begin{subfigure}{\shrinkfigtwo\textwidth}
        \includegraphics[width=\textwidth]{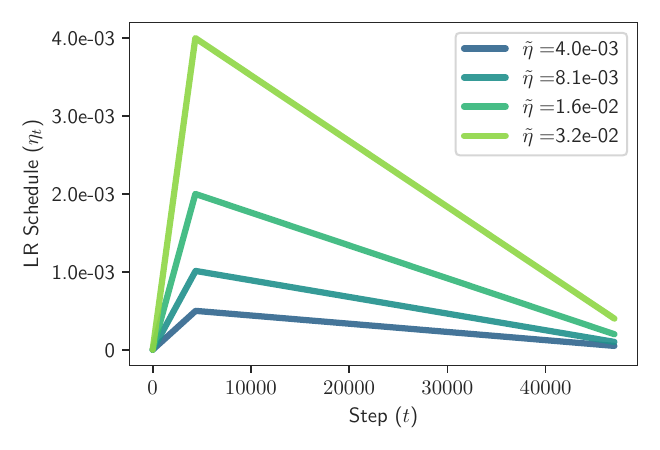}
      \end{subfigure}
      \hspace{-1mm}
      \begin{subfigure}{\shrinkfigtwo\textwidth}
        \includegraphics[width=\textwidth]{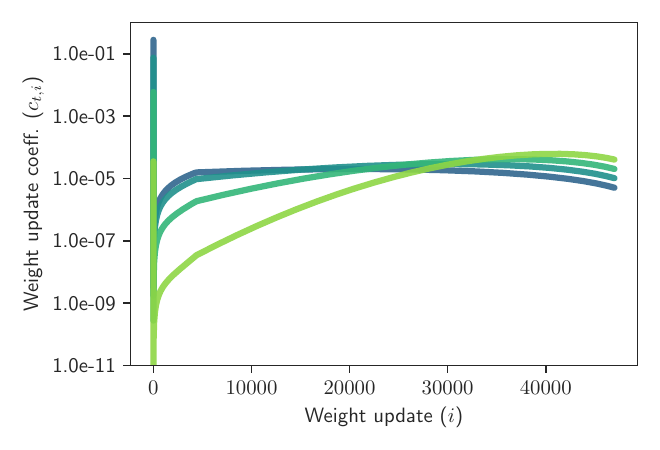}
      \end{subfigure}
      \vspace{-6.3mm}
      \subcaption{$\linear$ $\tenx$}
      \label{fig:inf_ema_lrs:tenx}
    \end{minipage}
  }
  \mbox{}
  \vspace{0.5mm}
  \mbox{}
  \makebox[\textwidth][c]{
    \begin{minipage}{\textwidth}
      \begin{subfigure}{\shrinkfigtwo\textwidth}
        \includegraphics[width=\textwidth]{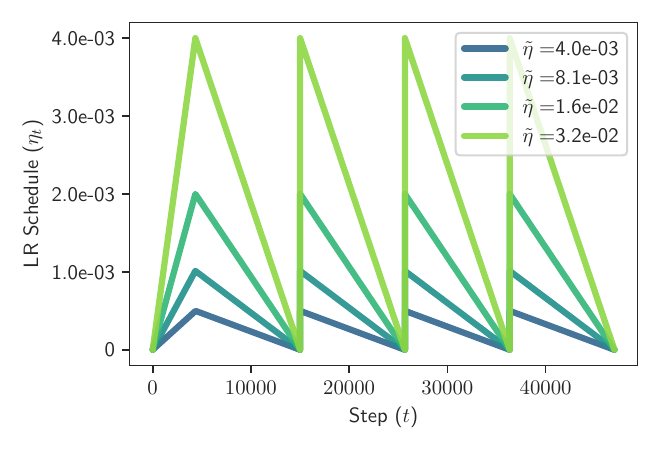}
      \end{subfigure}
      \hspace{-1mm}
      \begin{subfigure}{\shrinkfigtwo\textwidth}
        \includegraphics[width=\textwidth]{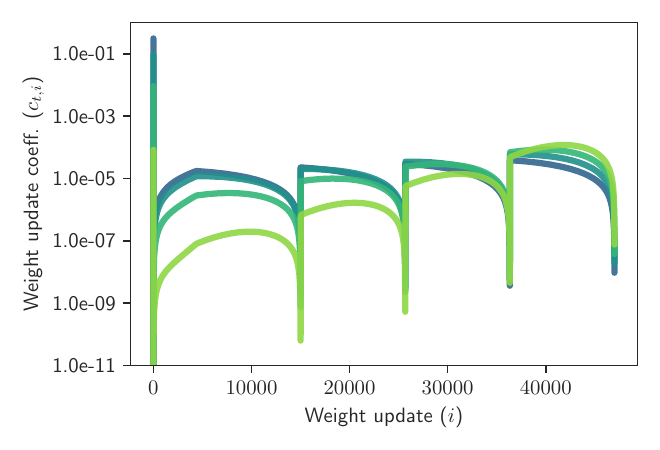}
      \end{subfigure}
      \vspace{-6.3mm}
      \subcaption{$\cyclic$}
      \label{fig:inf_ema_lrs:cyclic}
    \end{minipage}
  }
  \mbox{}
  \vspace{0.5mm}
  \mbox{}
  \makebox[\textwidth][c]{
    \begin{minipage}{\textwidth}
      \begin{subfigure}{\shrinkfigtwo\textwidth}
        \includegraphics[width=\textwidth]{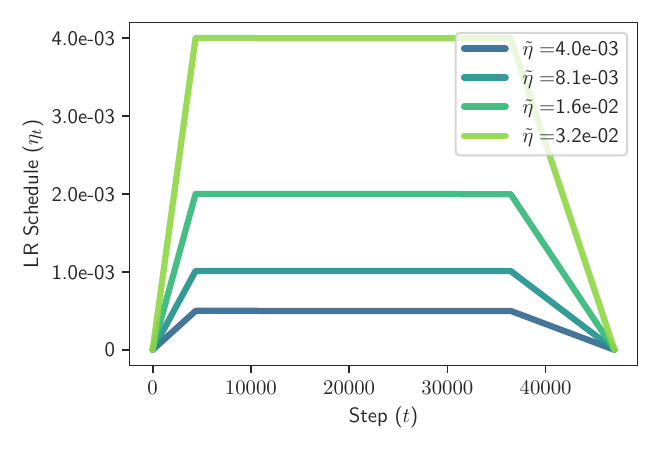}
      \end{subfigure}
      \hspace{-1mm}
      \begin{subfigure}{\shrinkfigtwo\textwidth}
        \includegraphics[width=\textwidth]{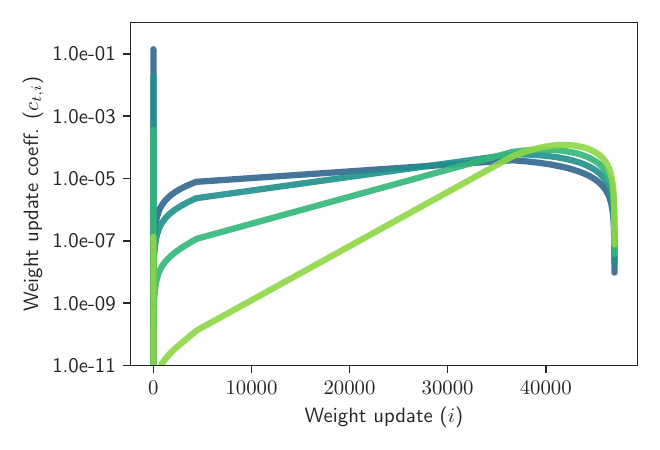}
      \end{subfigure}
      \vspace{-6.3mm}
      \subcaption{$\wsd$}
      \label{fig:inf_ema_lrs:wsd}
    \end{minipage}
  }
  \mbox{}
  \vspace{0.5mm}
  \mbox{}
  \caption{\textbf{LR curves and dual coefficients for standard
      vs.\ \emph{continuous} schedules}: Comparison of standard
    $\linear$ schedules versus $\cyclic$ and $\wsd$.  On the left are
    the exact LR schedules used in \cref{fig:maxLR_infinites}
    evaluations, i.e., 610M-parameter, 80~TPP training
    ($\rho$=$\nicefrac{1}{8}$, $\lambda$=$0.1$).  Dual coefficients
    shown at at final training step $t$=$47008$ (right side).  Note
    coefficient values are plotted on
    log-scale.\label{fig:inf_ema_lrs}}
\end{figure}

In this section, we provide some extra figures that were referenced in
the main paper.  \cref{fig:img_emas} shows the dual coefficients
at every step of training, using color to indicate the coefficient
value (log-scale).  Every horizontal row/step of
\cref{fig:img_emas} reflects the coefficients at that step,
essentially providing a version of \cref{fig:background_ema} but
at each step.
\cref{fig:more_ema_lrs} provides the LR schedules and dual
coefficients for some of the schedules discussed in the main paper, as
well as our proposed $\rational$ schedule, which combines all the
prior weight updates (after warmup) equally at every step.
Finally, \cref{fig:inf_ema_lrs} gives the LR schedules and dual
coefficients for the comparison to $\wsd$ and $\cyclic$ in
\cref{fig:maxLR_infinites}.

It is worth re-iterating that the dual coefficients can be computed
separately from any actual training.  They are mathematically
equivalent to the LR schedule itself and simply provide a perspective
on how the weight updates combine to form parameters, when using the
AdamW optimizer.
Furthermore, it is also worth noting the vertical bar at step 1 in the
dual coefficient plots; this bar reflects the coefficient on the
initial, random weights.
To some extent, this $c_{1,t}$ value can serve as an indicator of how
far the model has moved from initial conditions, i.e., the extent to
which the model has reduced the bias.  In general, if $c_{1,t}$ is too
high (e.g., when it outweighs the sum of the other coefficients), then
bias is likely significantly hindering learning.
Two effective ways to reduce $c_{1,t}$ and thus reduce bias are to
(1)~raise the peak LR, and (2)~train for more TPP\@.
In contrast to these two methods, raising weight decay, $\lambda$, can
also decrease $c_{1,t}$, but is counterproductive for reducing bias
because it also reduces the scale of weight updates, as noted in
\cref{hyp:wdbias}.  Likewise, using smaller batches also reduces
$c_{1,t}$, but is likewise counterproductive if the batches become too
small, to the extent that gradient noise increases.  However, $\dtoz$
is more robust to such noise than $\tenx$ or $\constant$ decay.
Further investigating the interplay of $\lambda$, batch size, and
peak learning rate is important future work.
% Make the point that WSD is unfair: -- UPDATE: this is not the point
%of the paper!  - It seems like WSD is unfair because if you *did*
%want intermediate checkpoints, say, 4 or 5 during the full run, from
%which you could make a scaling law, and you needed 20\% extra compute
%each time, well, you'd basically have to almost double the total
%amount of compute!  But cyclic doesn't have this!  - If you counted
%the decay part as part of your total budget, you'd end up training
%for fewer tokens.  So it's serious!  Cyclic has much more going for
%it.  - So is WSD-Cyclic actually better?

\end{document}